\documentclass[10pt,twocolumn,letterpaper]{article}

\usepackage[pagenumbers]{cvpr}    

\usepackage{stmaryrd}
\usepackage{amsthm}
\usepackage{booktabs}
\usepackage{graphicx}
\usepackage{placeins}
\usepackage{float}
\usepackage{tikz}
\usetikzlibrary{arrows.meta,positioning,calc,fit}

\definecolor{cvprblue}{rgb}{0.21,0.49,0.74}
\usepackage[
    pagebackref,
    breaklinks,
    colorlinks,
    allcolors=cvprblue
]{hyperref}

\hypersetup{
  pdftitle={Inference Equivalence and Quotient Coding},
  pdfauthor={Zibo Meng, Andrew C. M. Austin, Alin Achim}
}

\def\confName{}
\def\confYear{}

\title{Inference Equivalence and Quotient Coding}

\author{
Zibo Meng\textsuperscript{1} \quad
Andrew C. M. Austin\textsuperscript{1} \quad
Alin Achim\textsuperscript{1}\\[2pt]
\textsuperscript{1}\enspace University of Bristol
}

\newtheoremstyle{cvprplain}
  {4pt}
  {4pt}
  {\normalfont}
  {}
  {\normalfont}
  {.}
  {0.4em}
  {}

\theoremstyle{cvprplain}

\begin{document}

\maketitle


\begin{abstract}
Incomplete observations can admit multiple physical interpretations without making every task conclusion ambiguous. Recovering all compatible scenes is often unnecessary, whereas indiscriminate compression may change which claims are justified. We ask which distinctions among observation-compatible interpretation sets must be retained to preserve every valid conclusion of a specified task.
We first define inference validity as truth across all physically compatible states and characterize it, under a closedness condition, by distances to physically realizable falsifying measurements. For a finite semantic claim family ordered by specificity, the Exact Inference Frontier forms the unique minimal antichain generating all valid claims. We then define inference equivalence between compatibility sets and construct a canonical quotient code from their frontiers. Among deterministic codes admitting exact inference decoding, it induces the coarsest partition, minimizing the number of distinguishable code values rather than storage or computation.
Finally, we characterize inference errors from
learned compatibility sets, derive conditions for
exact inference preservation despite set errors,
and evaluate an asymmetric learning surrogate
under distribution shifts. On a registered finite
bank, 145 compatibility sets yield 31 quotient
codes with unchanged valid-claim sets. Across geometry,
mechanism, and joint shifts, learned estimates
exhibit different accuracy--specificity trade-offs.
On 1,962 official HRSID test images, 1,913
distinct candidate sets induce only 36 seven-claim
inference signatures, while a visual 3-NN baseline achieves 87.31\% precision on issued claims.
\end{abstract}


\section{Introduction}


Under occlusion, limited resolution, or missing measurements, the observed data may be compatible with several physical scene states. The realized scene can satisfy a claim that the observation does not justify, because a different compatible scene may falsify it. The central issue is therefore not merely whether a claim is factually true, but whether it holds across every interpretation consistent with the available measurements.


Reliable inference often precedes unique identification of the underlying scene. Explicitly representing every compatible physical interpretation can be costly, while compressing this uncertainty without regard to the task may eliminate a falsifying interpretation and introduce unsupported conclusions. Conversely, different compatibility sets may support exactly the same valid task claims, making their physical distinctions unnecessary for exact inference. This leads to the central question: which distinctions between observation-compatible interpretation sets must be retained to preserve all valid task inferences, and which can be eliminated without changing them?


Set-membership methods characterize states consistent with incomplete measurements \cite{1098995}, while possible-world semantics identifies conclusions shared across admissible interpretations \cite{10.1145/1634.1886}. Information bottleneck and computational mechanics provide principles for retaining task-relevant or predictive distinctions \cite{tishby2000informationbottleneckmethod,Shalizi_2001}. Incomplete multi-view classification uses view
completion \cite{Xu_2026_CVPR}, whereas conformal
methods quantify predictive uncertainty
\cite{3495724.3496026,11094306,pmlr-v180-mortier22a}. These studies establish important foundations for inference under incomplete information and equivalence-based representation. Building on these principles, we investigate how physically realizable counterexamples determine valid task conclusions, how inference-equivalent compatibility sets admit exact quotient coding, and how learned compatibility errors affect inference reliability.


Our contributions are threefold:

\par\noindent\textbf{\textit{i)}}\enspace
We characterize valid task claims across all states compatible with a physical observation, distinguishing observation-supported validity from factual truth. Under a closedness condition on physically realizable falsifying measurements, inference validity admits an exact distance-threshold characterization. For a finite claim family ordered by semantic
specificity, the Exact Inference Frontier is
the antichain of maximally specific valid claims;
its downward closure recovers every valid claim.
We establish refinement and local stability
properties and illustrate them in finite-bank
experiments.

\par\noindent\textbf{\textit{ii)}}\enspace
We identify which distinctions between
observation-compatible interpretation sets
affect valid task claims.
Exact Inference Frontiers induce a quotient
code preserving all valid task claims without
recovering the compatibility sets.
We prove that this code induces the coarsest
partition among deterministic
exact-inference-preserving encoders and
demonstrate the reduction in distinct code
values on finite banks.

\begin{samepage}
\par\noindent\textbf{\textit{iii)}}\enspace
We characterize inference errors caused by
estimated compatibility sets: omitted states can
introduce unsupported claims, while added states
can suppress valid ones.
We derive falsifier-witness conditions for exact
inference preservation even under two-sided
set errors.
We evaluate a closure-aware asymmetric learning
surrogate on a registered finite bank under
geometry, mechanism, and joint shifts,
measuring unsupported claims, lost valid
claims, and supported specificity.
\par
\end{samepage}


\section{Related Work}

\subsection{Inference under Incomplete Information}

Set-membership methods characterize states
consistent with perturbed measurements
\cite{1098995}; possible-world semantics
describes admissible completions
\cite{10.1145/1634.1886}, and certain-answer
theory formalizes conclusions shared
across them \cite{3031929.3031970}.
These foundations motivate inference validity
but do not directly provide physically
realizable falsification margins or maximal
valid claims under task-specific specificity.
Xu et al.\ use external visual knowledge
to guide missing-view completion for
classification \cite{Xu_2026_CVPR}.
We characterize valid claims through
realizable falsifiers and construct their
Exact Inference Frontier.

\subsection{Information Equivalence and Compression}

The information bottleneck retains
target-relevant information under compression
\cite{tishby2000informationbottleneckmethod};
computational mechanics identifies histories
with identical predictive distributions
\cite{Shalizi_2001}, while MDP abstraction
preserves specified decision properties
\cite{LiWalshLittman2006}.
Incomplete-database theory also studies
query-relative equivalence of possible-world
representations with identical certain answers
\cite{10.1145/1634.1886}.
Thus, equivalence-based quotienting is not
itself a new compression principle.

Our contribution is to characterize the
physically constrained, task-dependent
inference map. Under the stated conditions,
realizable falsifiers determine valid claims,
semantic specificity defines their frontier,
and the frontier induces the coarsest
exact-inference-preserving deterministic code.
Its optimality concerns distinct code values,
not storage or computational complexity.
We distinguish finite-bank preservation
from full-model validity and analyze
inference errors from learned compatibility sets.

\subsection{Reliable Inference}

Abstract interpretation emphasizes sound
over-approximation \cite{CousotCousot1977}.
Selective visual question answering uses
estimated reliability for abstention
\cite{KhanFu2024}, while counterfactual
evidence dropout probes visual support
\cite{Kotte2026EVICT}.
Conformal prediction provides marginal
coverage under exchangeability
\cite{3495724.3496026}, with structured
extensions to hierarchical outputs
\cite{pmlr-v180-mortier22a}.
These approaches provide soundness or
predictive-reliability guarantees in their
respective settings, but do not directly
characterize the exact frontier of task claims
supported by a given physical observation.

We distinguish unsupported claims caused
by omitted compatible falsifiers from
valid claims suppressed by spurious
falsifiers in estimated compatibility sets.
We evaluate both errors and supported
specificity under geometry, mechanism,
and joint distribution shifts.


\section{Methodology}


\subsection{Inference Validity and Exact Frontier}
\label{sec:inference_validity}

Let $s^\star\in\mathcal S_{\mathrm{phys}}$ be the unknown
physical state, where $\mathcal S_{\mathrm{phys}}$ contains
all physically admissible states. An observation
$x\in\mathcal X\subseteq\mathbb R^d$ is generated by
\begin{equation}
x=\mathcal O_{\eta^\star}(s^\star)+e,
\qquad \|e\|\le\epsilon,
\label{eq:physical_observation}
\end{equation}
Here $\mathcal O_\eta$ is the physical observation operator,
$\eta^\star$ is the acquisition configuration, and
$\epsilon\geq 0$ bounds measurement error. Given
metadata $m$, let $\mathcal H(m)$ denote the admissible
acquisition configurations. The compatible-state set is
\begin{equation}
\Gamma_\epsilon(x;m)=
\left\{
s\in\mathcal S_{\mathrm{phys}}:
\begin{array}{l}
\exists\eta\in\mathcal H(m),\\
\|\mathcal O_\eta(s)-x\|\le\epsilon
\end{array}
\right\}.
\label{eq:physical_compatibility}
\end{equation}
Assuming a correct physical model and
$\eta^\star\in\mathcal H(m)$, we have
$s^\star\in\Gamma_\epsilon(x;m)\neq\varnothing$.
An empty compatibility set indicates
model--observation inconsistency.

Let $\mathcal Q_\tau$ be a finite family of binary task
claims $q:\mathcal S_{\mathrm{phys}}\to\{0,1\}$.
Writing $T_q=\{s\in\mathcal S_{\mathrm{phys}}:q(s)=1\}$,
we identify claims with identical truth sets and define
\begin{equation}
V_\tau(\Gamma)
=\{q\in\mathcal Q_\tau:\Gamma\subseteq T_q\}.
\label{eq:invariant_inference}
\end{equation}
A claim is valid only if it holds for every compatible
state; $q(s^\star)=1$ alone does not imply
$q\in V_\tau(\Gamma_\epsilon(x;m))$.

For each claim $q$, define its physically realizable
falsifying measurements under metadata $m$ by
\begin{equation}
\mathcal M_q^-(m)=
\left\{
\mathcal O_\eta(s):
\begin{array}{l}
s\in\mathcal S_{\mathrm{phys}},\ q(s)=0,\\
\eta\in\mathcal H(m)
\end{array}
\right\}.
\label{eq:falsification_set}
\end{equation}
The falsification margin is the distance from
$x$ to this set:
\begin{equation}
\rho(q;x,m)
=
\inf_{y\in\mathcal M_q^-(m)}
\|x-y\|.
\label{eq:falsification_margin}
\end{equation}
We set $\rho(q;x,m)=+\infty$ when
$\mathcal M_q^-(m)=\varnothing$. If $\mathcal M_q^-(m)$ is closed, validity admits
the exact measurement-space characterization
\begin{equation}
q\in V_\tau(\Gamma_\epsilon(x;m))
\iff \rho(q;x,m)>\epsilon.
\label{eq:physical_validity_certificate}
\end{equation}
A claim is invalid exactly when an admissible falsifying
measurement lies within $\epsilon$ of $x$.
For a nonempty closed falsification set, the minimum
distance is attained. Closedness follows, for example,
if the admissible falsifying state--configuration domain
is compact and $(s,\eta)\mapsto\mathcal O_\eta(s)$
is continuous.

For a finite registered bank
$\mathcal B\subseteq
\mathcal S_{\mathrm{phys}}\times\mathcal H(m)$
of admissible state--configuration pairs, let
$\mathcal M^-_{q,\mathcal B}(m)\subseteq\mathcal M^-_q(m)$
denote its falsifying measurements and
$\rho_{\mathcal B}$ their distance from $x$, set to
$+\infty$ when the set is empty.
A bank counterexample within $\epsilon$ certifies
invalidity in the full physical model.
If every physical falsifying measurement lies within
a certified distance $\delta_q<\infty$ of some bank
falsifying measurement, then
\begin{equation}
\rho_{\mathcal B}-\delta_q
\le\rho\le\rho_{\mathcal B}.
\label{eq:certified_margin_bounds}
\end{equation}
Here the arguments $(q;x,m)$ are suppressed.
Thus, $\rho_{\mathcal B}>\epsilon+\delta_q$
certifies validity over the full physical model.
Without a certified covering bound, bank-based
validity is established only relative to the
registered physical realizations.

Define the semantic specificity order by
$q_1\preceq_\tau q_2$ iff
$T_{q_2}\subseteq T_{q_1}$,
so $q_2$ is at least as specific as $q_1$.
With identical truth sets identified as above,
$\preceq_\tau$ is a partial order.
The Exact Inference Frontier is
\begin{equation}
F_\tau(\Gamma)
=
\operatorname{Max}_{\preceq_\tau}
V_\tau(\Gamma).
\label{eq:exact_inference_frontier}
\end{equation}
If $\mathcal M^-_{q'}(m)$ is closed for every
$q'\in\mathcal Q_\tau$, validity and maximality
yield the exact frontier criterion:
\begin{equation}
\begin{aligned}
q\in F_\tau(\Gamma_\epsilon(x;m))
\iff {}& \rho(q;x,m)>\epsilon,\\
& \forall q'\succ_\tau q:\ 
\rho(q';x,m)\le\epsilon.
\end{aligned}
\label{eq:frontier_certificate}
\end{equation}
Since $\mathcal Q_\tau$ is finite and
$V_\tau(\Gamma)$ is downward closed under
$\preceq_\tau$, we have
\begin{equation}
V_\tau(\Gamma)=\downarrow F_\tau(\Gamma),
\label{eq:frontier_completeness}
\end{equation}
where $\downarrow F=\{q\in\mathcal Q_\tau:
\exists f\in F,\ q\preceq_\tau f\}$.
Every generating subset must contain all maximal
valid claims. Thus, $F_\tau(\Gamma)$ is the
unique inclusion-minimal generating antichain
of $V_\tau(\Gamma)$.

For nonempty compatible sets
$\Gamma_2\subseteq\Gamma_1$, observation
refinement preserves all previously valid claims:
\begin{equation}
V_\tau(\Gamma_1)
\subseteq V_\tau(\Gamma_2),
\label{eq:inference_refinement}
\end{equation}
Moreover, each $f_1\in F_\tau(\Gamma_1)$
satisfies $f_1\preceq_\tau f_2$ for some
$f_2\in F_\tau(\Gamma_2)$.
Thus, refinement can yield more specific valid
claims, although the frontiers themselves
need not be nested.

For fixed metadata $m$, the falsification set
$\mathcal M_q^-(m)$ is independent of $x$.
When this set is nonempty, its distance function
is $1$-Lipschitz:
$|\rho(q;x,m)-\rho(q;x',m)|\le\|x-x'\|$.
Define the validity-threshold separation
\begin{equation}
\Delta_\tau(x;m)=
\min_{\substack{
q\in\mathcal Q_\tau\\
\mathcal M_q^-(m)\neq\varnothing
}}
|\rho(q;x,m)-\epsilon|,
\label{eq:frontier_stability_radius}
\end{equation}
where $\min\varnothing=+\infty$.
For fixed metadata $m$ and tolerance $\epsilon$,
if $\Delta_\tau(x;m)>0$ and
$\|x'-x\|<\Delta_\tau(x;m)$, then for
model-consistent observations,
\begin{equation}
\begin{aligned}
V_\tau(\Gamma_\epsilon(x';m))
&=V_\tau(\Gamma_\epsilon(x;m)),\\
F_\tau(\Gamma_\epsilon(x';m))
&=F_\tau(\Gamma_\epsilon(x;m)).
\end{aligned}
\label{eq:local_frontier_stability}
\end{equation}
No validity threshold is crossed under such a
perturbation, while claims with empty falsification
sets remain valid.


\subsection{Quotient Coding}
\label{sec:quotient_coding}

Let $\mathcal A$ denote the admissible
observation--metadata--tolerance triples
$(x,m,\epsilon)$, and let $\mathcal G$
be the family of nonempty compatibility sets
$\Gamma_\epsilon(x;m)$ induced by
$(x,m,\epsilon)\in\mathcal A$.
We seek a deterministic representation
$r:\mathcal G\to\mathcal Z$ and an exact
inference decoder $d:\mathcal Z\to
2^{\mathcal Q_\tau}$ that minimize the number
of distinguishable code values:
\begin{equation}
\min_{r,d}|\operatorname{im}r|
\quad\text{s.t.}\quad
d\circ r=V_\tau\ \text{on }\mathcal G.
\label{eq:quotient_objective}
\end{equation}
This objective concerns code cardinality,
not storage length, representation dimension,
or computational cost.

For nonempty $\Gamma\in\mathcal G$, define
its binary inference signature
$\sigma_\tau(\Gamma)=
(\min_{s\in\Gamma}q(s))_{q\in\mathcal Q_\tau}$.
Two compatibility sets are inference-equivalent
exactly when they support the same valid
task claims:
\begin{equation}
\begin{aligned}
\Gamma_1\sim_\tau\Gamma_2
&\iff V_\tau(\Gamma_1)=V_\tau(\Gamma_2)\\
&\iff \sigma_\tau(\Gamma_1)
=\sigma_\tau(\Gamma_2).
\end{aligned}
\label{eq:inference_equivalence}
\end{equation}
Each signature component indicates whether
its claim holds across all compatible states.
Thus, inference equivalence does not require
identical physical compatibility sets.

\begin{figure*}[!t]
    \centering
    \includegraphics[
        width=0.76\textwidth
    ]{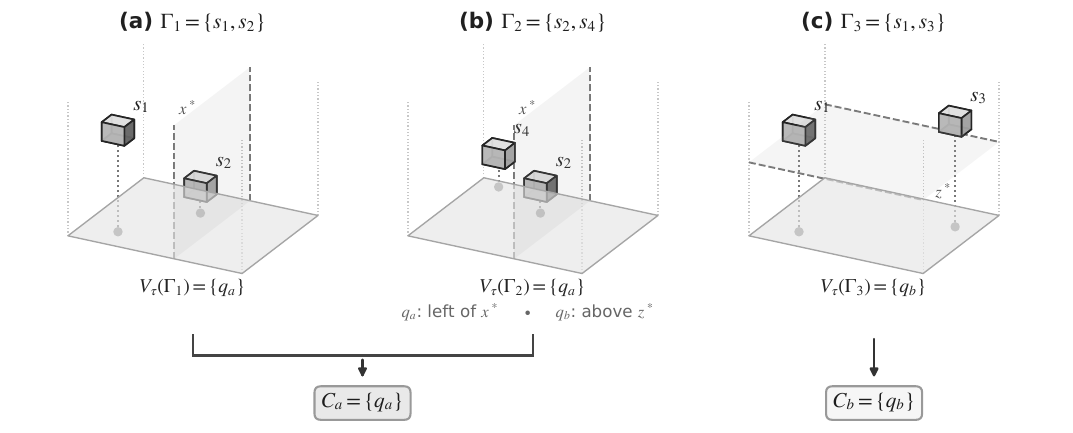}
    \caption{
    \textbf{Inference equivalence and quotient coding.}
    (a,b) Distinct compatibility sets
    $\Gamma_1\neq\Gamma_2$ support identical
    valid task claims and therefore share
    the quotient code $C_a$.
    (c) A set supporting different valid
    claims requires a distinct code $C_b$.
    The configurations are illustrative,
    not experimental results.
}
    \label{fig:inference_equivalence}
\end{figure*}

Figure~\ref{fig:inference_equivalence} illustrates
the distinction between physical recovery and
exact task inference.
Although $\Gamma_1=\{s_1,s_2\}$ and
$\Gamma_2=\{s_2,s_4\}$ contain different
physical possibilities, they support
exactly the same valid claims,
$V_\tau(\Gamma_1)=V_\tau(\Gamma_2)=\{q_a\}$,
and therefore share the quotient code $C_a$.
In contrast, $\Gamma_3=\{s_1,s_3\}$ supports
$\{q_b\}$ and requires a distinct code $C_b$.
Thus, exact inference preservation does not
require recovering the compatibility set itself;
only distinctions that change valid task claims
must be retained.
Our contribution formalizes this principle
under physical observation compatibility
and uses the Exact Inference Frontier to
construct the coarsest deterministic
exact-inference-preserving quotient
(Eqs.~\ref{eq:inference_equivalence}
and~\ref{eq:quotient_code}).

The task closure collects all physical states
satisfying every claim valid on $\Gamma$:
\begin{equation}
K_\tau(\Gamma)=
\bigcap_{q\in V_\tau(\Gamma)}T_q,
\label{eq:task_closure}
\end{equation}
where the empty intersection denotes
$\mathcal S_{\mathrm{phys}}$. It satisfies
\begin{equation}
\Gamma\subseteq K_\tau(\Gamma),\qquad
V_\tau(K_\tau(\Gamma))=V_\tau(\Gamma).
\label{eq:task_closure_properties}
\end{equation}
Consequently, $K_\tau$ is idempotent, and
$\Gamma_1\sim_\tau\Gamma_2$ if and only if
$K_\tau(\Gamma_1)=K_\tau(\Gamma_2)$.
Moreover, $K_\tau(\Gamma)$ is the largest
physical-state subset preserving
$V_\tau(\Gamma)$, but need not be realizable
as an observation-compatible set
$\Gamma_\epsilon(x;m)$.

Inference equivalence induces the quotient
space $\mathcal G/{\sim_\tau}$.
Using the Exact Inference Frontier, define
the canonical quotient code and decoder by
\begin{equation}
\begin{aligned}
C_\tau(\Gamma)&=F_\tau(\Gamma),\\
D_\tau(z)&=\downarrow_\tau z,\\
D_\tau(C_\tau(\Gamma))&=V_\tau(\Gamma).
\end{aligned}
\label{eq:quotient_code}
\end{equation}
Here $z\in\operatorname{im}C_\tau$ and
$\downarrow_\tau z$ contains all task claims
implied by some $f\in z$.
Frontier completeness gives
$\Gamma_1\sim_\tau\Gamma_2$ if and only if
$C_\tau(\Gamma_1)=C_\tau(\Gamma_2)$.
Thus, the code preserves every valid task
claim without retaining the full
compatibility set.

Under the closed-falsification-set
condition of Sec.~\ref{sec:inference_validity},
the canonical code consists exactly of
the $\preceq_\tau$-maximal task claims
satisfying $\rho(q;x,m)>\epsilon$.
For a finite physical bank
$\mathcal B_m\subseteq
\mathcal S_{\mathrm{phys}}\times\mathcal H(m)$,
define the bank-induced compatibility set
\begin{equation}
\label{eq:bank_compatibility}
\begin{aligned}
\Gamma_{\epsilon,\mathcal B}(x;m)
=\big\{s:\;&\exists\eta,\ (s,\eta)\in\mathcal B_m,\\
&\|\mathcal O_\eta(s)-x\|\leq\epsilon
\big\}.
\end{aligned}
\end{equation}
For the registered pairs
$(s,\eta)\in\mathcal B_m$, the
bank-restricted falsification set
$\mathcal M^-_{q,\mathcal B}(m)$ contains
the measurements $\mathcal O_\eta(s)$
with $q(s)=0$.
Let $\rho_{\mathcal B}(q;x,m)$ denote
the distance from $x$ to this set,
with value $+\infty$ when it is empty.
Since $\mathcal B_m$ is finite, the
falsification set is closed. For nonempty
$\Gamma_{\epsilon,\mathcal B}(x;m)$,
the canonical code is therefore
\begin{equation}
C_\tau(\Gamma_{\epsilon,\mathcal B}(x;m))
=\operatorname{Max}_{\preceq_\tau}
\{q\in\mathcal Q_\tau:
\rho_{\mathcal B}(q;x,m)>\epsilon\}.
\label{eq:bank_quotient}
\end{equation}
This characterization is exact only for
the registered bank. Extending a
bank-derived validity claim to the full
physical model requires an additional
certificate, such as the coverage bound
in Sec.~\ref{sec:inference_validity}.

Since
$\Gamma_{\epsilon,\mathcal B}(x;m)
\subseteq\Gamma_\epsilon(x;m)$,
for nonempty compatibility sets,
\begin{equation}
V_\tau(\Gamma_\epsilon(x;m))
\subseteq
V_\tau(\Gamma_{\epsilon,\mathcal B}(x;m)).
\label{eq:bank_full_inference_inclusion}
\end{equation}
Thus, the bank and full-model quotient codes
coincide if and only if every bank-valid
claim is also full-model valid.
Under the coverage bound of
Sec.~\ref{sec:inference_validity},
a sufficient certificate is
$\rho_{\mathcal B}(q;x,m)>\epsilon+\delta_q$
for every bank-valid claim $q$.
Failure to certify a claim does not
imply its invalidity.

We now characterize the coarsest exact
deterministic code. Let
$r:\mathcal G\to\mathcal Z$ and
$d:\mathcal Z\to 2^{\mathcal Q_\tau}$
satisfy $d\circ r=V_\tau$ on $\mathcal G$.
For $z\in\operatorname{im}r$, define
$\phi_\tau(z)=
\operatorname{Max}_{\preceq_\tau}d(z)$.
Frontier completeness implies
$C_\tau=\phi_\tau\circ r$.
Thus, every exact code must distinguish
different inference-equivalence classes.
Since $D_\tau\circ C_\tau=V_\tau$,
this lower bound is attained:
\begin{equation}
\begin{aligned}
\min_{r,d}|\operatorname{im}r|
&=|\mathcal G/{\sim_\tau}|\\
&=|\operatorname{im}C_\tau|
=|\operatorname{im}V_\tau|
\le 2^{|\mathcal Q_\tau|}.
\end{aligned}
\label{eq:quotient_minimality}
\end{equation}
Here the minimum is over deterministic
encoders and exact decoders satisfying
the stated constraint. Every optimizer
induces the inference-equivalence partition,
up to bijective relabeling of code values.
This establishes minimum code cardinality,
not minimum storage length, representation
dimension, or computational cost.

Finally, quotient coding is task-relative.
Let $\mathcal Q_\tau\subseteq
\mathcal Q_{\tau^+}$, with unchanged
truth sets and specificity order on
the original claims. Then
$V_\tau(\Gamma)=
V_{\tau^+}(\Gamma)\cap\mathcal Q_\tau$,
so $\sim_{\tau^+}\subseteq\sim_\tau$.
Define
$\pi_{\tau\leftarrow\tau^+}(z)=
\operatorname{Max}_{\preceq_\tau}
(\downarrow_{\tau^+}z\cap\mathcal Q_\tau)$.
Frontier completeness gives
\begin{equation}
C_\tau=
\pi_{\tau\leftarrow\tau^+}\circ C_{\tau^+}.
\label{eq:quotient_refinement}
\end{equation}
Thus, adding task claims can split
inference-equivalence classes but cannot
merge previously distinct classes.

The task closure also characterizes
inference-preserving enlargements of
compatibility sets.
For $\Gamma\subseteq\widehat{\Gamma}$,
exact task inference is preserved if and only if
$\widehat{\Gamma}\subseteq K_\tau(\Gamma)$.
Thus, exact inference does not require exact
compatibility-set recovery. This observation
motivates the compatibility-learning analysis
in Sec.~3.3.


\subsection{Compatibility Learning and Reliable Inference}
\label{sec:compatibility_learning}

We study when errors in learned compatibility
sets change valid task claims and when
exact inference is preserved.

Let $\mathcal B\subseteq\mathcal S_{\mathrm{phys}}$
be a finite registered state bank.
For observation $o=(y,M,h,\varepsilon)$,
specifying measurements, visibility mask,
intervention history, and tolerance, define
the bank-relative compatibility set
\begin{equation}
\label{eq:sc33_compatibility}
\Gamma_{\mathcal B}(o)=
\{s\in\mathcal B:
d_{M,h}(y,\mathcal O_h(s))\le\varepsilon\}.
\end{equation}

Here $\mathcal O_h(s)$ is the registered measurement
under history $h$, and $d_{M,h}$ is the masked
observation distance. Unlike the full compatibility
set in Sec.~\ref{sec:inference_validity},
$\Gamma_{\mathcal B}(o)$ is restricted to
$\mathcal B$. We analyze only observations with
$\Gamma_{\mathcal B}(o)\neq\varnothing$;
empty-set cases are excluded.

For each $s\in\mathcal B$, a candidate-wise
learner produces a compatibility score
$p_\theta(s\mid o,\mathcal B)
=f_\theta(\phi(o,\mathcal O_h(s)))$,
where $\phi$ extracts observation--template features.
The estimated set is

\begin{equation}
\widehat\Gamma(o)
=
\{s\in\mathcal B:
p_\theta(s\mid o,\mathcal B)\geq t\},
\label{eq:sc33_prediction}
\end{equation}

The threshold $t$ is selected on validation data;
multiple candidates may be retained.
Write $\Gamma_o=\Gamma_{\mathcal B}(o)$ and
$\widehat{\Gamma}_o=\widehat{\Gamma}(o)$.

For $q\in\mathcal Q_\tau$, let
$E_q=T_q\cap\mathcal B$.
For nonempty $C\subseteq\mathcal B$, write
$V(C)=V_{\tau,\mathcal B}(C)
=\{q\in\mathcal Q_\tau:C\subseteq E_q\}$,
$F_{\tau,\mathcal B}(C)
=\operatorname{Max}_{\preceq_\tau}V(C)$,
and $K(C)=\bigcap_{q\in V(C)}E_q$,
with an empty intersection equal to $\mathcal B$.
As in Sec.~\ref{sec:quotient_coding},
$C\subseteq K(C)$ and $V(K(C))=V(C)$.

We quantify compatibility-set errors using
the task-relative discrepancy
\begin{equation}
\label{eq:sc33_discrepancy}
\begin{aligned}
\mathcal E_{\tau,\mathcal B}
&=|\Gamma_o\setminus\widehat{\Gamma}_o|\\
&\quad+\lambda
|\widehat{\Gamma}_o\setminus K(\Gamma_o)|,
\quad \lambda>0.
\end{aligned}
\end{equation}
The first term penalizes omitted compatible states
that may introduce unsupported claims; the second
penalizes added states outside the task closure,
which invalidate previously valid claims.
The discrepancy vanishes exactly when
$\Gamma_o\subseteq\widehat{\Gamma}_o
\subseteq K(\Gamma_o)$, guaranteeing
$V(\widehat{\Gamma}_o)=V(\Gamma_o)$.
Exact inference can remain unchanged even when
this discrepancy is positive.

These asymmetric errors distinguish
task-level inference reliability from
exact compatibility-set recovery.
Bank-relative validity extends to the full
physical model only under additional
certified coverage conditions
(Sec.~\ref{sec:inference_validity}).
Further geometric illustrations are
provided in Appendix~\ref{app:specificity}.

We study task-relative compatibility learning
through candidate-wise classification, comparing
symmetric, coverage-weighted, closure-target,
and corridor-oriented asymmetric objectives. For corridor-oriented learning, the candidate-wise
label is $\mathbf{1}\{s\in K(\Gamma_o)\}$.
We train an ExtraTrees classifier with weighted
Gini splitting, assigning weights $(12,0.1,3)$
to $\Gamma_o$, $K(\Gamma_o)\setminus\Gamma_o$,
and $\mathcal B\setminus K(\Gamma_o)$,
respectively. This objective is a surrogate for
Eq.~\ref{eq:sc33_discrepancy},
not its direct minimization.

For nonempty $\widehat{\Gamma}_o$, the
predicted inference set is
$\widehat V(o)=V(\widehat{\Gamma}_o)$.
If $\widehat{\Gamma}_o=\varnothing$,
we instead set $\widehat V(o)=\varnothing$
(abstention), avoiding vacuous inference.
The closure-based guarantees below concern
nonempty predicted compatibility sets.

For nonempty $\widehat{\Gamma}_o$,
$\Gamma_o\subseteq\widehat{\Gamma}_o$
implies
$\widehat V(o)\subseteq V(\Gamma_o)$,
preventing unsupported bank-relative claims.
Conversely,
$\widehat{\Gamma}_o\subseteq K(\Gamma_o)$
implies
$V(\Gamma_o)\subseteq\widehat V(o)$,
preserving all originally valid claims.
Together, these inclusions give the
exact criterion below.

For every estimate satisfying
$\varnothing\neq\Gamma_o
\subseteq\widehat{\Gamma}_o\subseteq\mathcal B$,
\begin{equation}
\label{eq:sc33_exact_preservation}
\widehat V(o)=V(\Gamma_o)
\iff
\widehat{\Gamma}_o\subseteq K(\Gamma_o).
\end{equation}

Equation~\eqref{eq:sc33_exact_preservation}
gives a necessary and sufficient
condition for exact inference preservation
under $\Gamma_o\subseteq\widehat{\Gamma}_o$.
Without this assumption, the corridor remains
sufficient but not necessary: an estimate may
omit compatible states while preserving every
valid task claim.

We evaluate bank-relative coverage and
normalized unsupported-inference risk:

\begin{equation}
\begin{aligned}
\mathrm{Cov}
&=\Pr_o[\Gamma_o\subseteq\widehat\Gamma_o],
\\
R_{\mathrm{unsup}}
&=\mathbb E_o
\left[
\frac{|\widehat V(o)\setminus V(\Gamma_o)|}
{|\mathcal Q_\tau|}
\right].
\end{aligned}
\label{eq:sc33_reliability}
\end{equation}

On the event
$\Gamma_o\subseteq\widehat{\Gamma}_o$,
monotonicity gives
$\widehat V(o)\subseteq V(\Gamma_o)$.
Since the normalized loss is at most one,
$R_{\mathrm{unsup}}\le 1-\mathrm{Cov}$.
This inequality holds for any evaluation
distribution but does not guarantee high
coverage under distribution shift.

We also report the exact-preservation rate
$P_{\mathrm{exact}}
=\Pr_o[\widehat V(o)=V(\Gamma_o)]$
and normalized lost-claim risk
$R_{\mathrm{lost}}
=\mathbb E_o[
|V(\Gamma_o)\setminus\widehat V(o)|
/|\mathcal Q_\tau|]$.
Together with $R_{\mathrm{unsup}}$,
these distinguish lost valid claims from
unsupported predictions.

Unsupported claims have a bank-relative
falsifier characterization.
For $q\in\mathcal Q_\tau$, define
$Z_q=\mathcal B\setminus E_q$.
When $\widehat{\Gamma}_o\neq\varnothing$,

\begin{equation}
\begin{aligned}
q\in\widehat V(o)\setminus V(\Gamma_o)
\iff{}&
\widehat\Gamma_o\cap Z_q=\varnothing
\\[-1pt]
&\land\ \Gamma_o\cap Z_q\neq\varnothing.
\end{aligned}
\label{eq:sc33_falsifiers}
\end{equation}

Thus, an unsupported claim has a falsifier
in $\Gamma_o$ but none in
$\widehat{\Gamma}_o$.

For claim-level analysis, define the
bank-relative specificity score
$\operatorname{Spec}_{\mathcal B}(q)
=1-|E_q|/|\mathcal B|$.
If $T_{q_2}\subseteq T_{q_1}$, then
$\operatorname{Spec}_{\mathcal B}(q_2)
\geq\operatorname{Spec}_{\mathcal B}(q_1)$.
This measures exclusion within the
registered bank, not absolute semantic depth.

We learn bank-relative compatibility from registered
RGB measurements using history-gated candidate-pair
features. Models, features, thresholds, and evaluation
conditions are fixed before prospective testing on
unseen registered geometries, masking mechanisms,
and their joint shift. Candidate templates remain
available under the physical-bank protocol.
These finite-bank results do not establish
continuous-state recovery or reliability for
unregistered physical states or observation operators.


\section{Experiments}


\subsection{Ambiguity and Inference Frontiers}
\label{sec:observation_ambiguity}

Using Push-T-based simulations
\cite{10.1177/02783649241273668} with occluded
views and partial temporal observations, we test
whether ambiguous measurements still support valid
task claims, how ambiguity affects claim specificity,
and how additional observations refine inference
frontiers.

We evaluate 13 registered claims across
5,760 historical and 3,040 prospective
observation contexts, spanning geometry,
visibility-mechanism, and joint shifts. Let
$o_i=(y_i,M_i,h_i,\epsilon_i)$ and
$\Gamma_i=\Gamma_{\mathcal B}(o_i)$.
Independent measurement-space falsification
margins and registered truth-set checks agree
on bank-validity for every observation.
All oracle compatibility sets are nonempty
and contain their generating states.
These checks validate the finite-bank
implementation, not validity over the
continuous state space.
The margin definition and temporal witness
are provided in Appendix~\ref{app:temporal_refinement}.

For each observation, let
$V_i=V_{\tau,\mathcal B}(\Gamma_i)$
and
$F_i=\operatorname{Max}_{\preceq_\tau}V_i$.
All frontiers satisfy
$V_i=\downarrow F_i$, with no antichain
or minimality violations.

We define $P_{\mathrm{partial}}$ as the
fraction of observations satisfying
$|\Gamma_i|>1$ and
$V_i\cap\mathcal Q_{\tau,\mathcal B}^{\circ}
\neq\varnothing$,
where $\mathcal Q_{\tau,\mathcal B}^{\circ}$
contains nontrivial bank-relative claims.
Thus, Partial-ID measures informative
task inference despite physical ambiguity.

Across 5,760 historical observations,
73.85\% admit multiple compatible states,
while 49.38\% are both ambiguous and
support nontrivial valid claims.
This pattern persists across prospective
shifts. Multiple incomparable maximal claims occur
in 56.39\% of historical contexts.
Even when $|\Gamma_i|=2$, 12 distinct
frontiers occur. Thus, compatibility-set
cardinality alone does not determine
task-relevant information.
Full regime-wise statistics are reported in
Appendix~\ref{app:observation_ambiguity_statistics},
and physical-state witnesses are illustrated
in Fig.~\ref{fig:physical_inference_equivalence}.

A claim may be true at the generating state
but unsupported by the observation.
Using the known generating state $s_i^\star$,
we define $L_{\mathrm{CU}}$ as the fraction
of true claim instances $q(s_i^\star)=1$
for which $q\notin V_i$.
The true-but-unsupported rate is $0.3713$
historically and $0.2816$--$0.3104$
under prospective shifts. These rates quantify the gap between factual truth
and bank-relative observational support, not
classifier error: a claim true at the generating
state is unsupported whenever another compatible
bank state falsifies it.

Changing spatial visibility altered the frontier
in 40.1\% of 4,320 controlled comparisons,
whereas 718 pairs with different compatibility
sets shared the same frontier. Nested temporal
observations could exclude falsifiers and support
new claims without invalidating previously valid
ones. Across 1,556 evaluated positive-distance
pairs below the finite-bank stability threshold,
frontiers remained unchanged. 
These controls
separate physical ambiguity from task-relevant
information; detailed counts and witnesses
appear in Appendix~\ref{app:temporal_refinement}.

\begin{figure*}[t]
    \centering
    \includegraphics[width=0.90\textwidth]
    {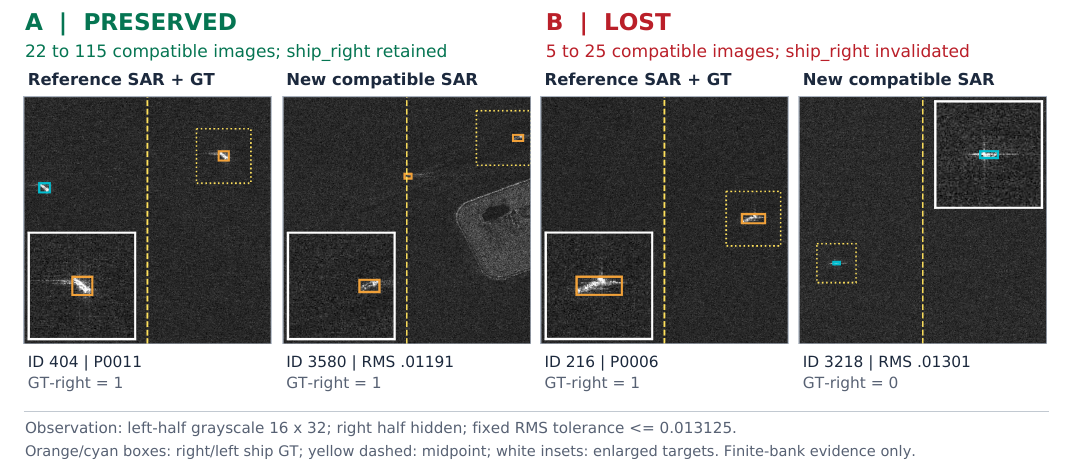}

   \caption{
\textbf{Real-SAR inference preservation and loss.}
Under the same frozen observation operator,
RMS tolerance, and nested HRSID training-image
banks, \textbf{(A)} compatibility expansion
from 22 to 115 images preserves
\texttt{ship\_right}, whereas
\textbf{(B)} expansion from 5 to 25 admits
a compatible falsifier and invalidates the claim.
Boxes indicate ground-truth ship annotations;
insets magnify selected targets.
}

    \label{fig:real_sar_inference}
\end{figure*}

To examine the same inference distinction in real SAR imagery, we evaluate
HRSID training images \cite{9127939}
under a frozen left-half
grayscale observation ($16\times32$) and nested
image banks ranging from 640 to 3,642 images.
Among 22 initially source-diverse anchors
supporting \texttt{ship\_right}, nine retain
the claim in the expanded bank, whereas
13 lose it.
Figure~\ref{fig:real_sar_inference} contrasts
two real-SAR witnesses: compatibility expansion
preserves \texttt{ship\_right} when no new
falsifier enters, but invalidates the claim
when a compatible falsifier appears.
Thus, inference reliability depends on which
explanations remain possible, not merely
how many.
These development-only HRSID results establish
the distinction within frozen observational
image banks, without implying full physical-SAR
validity or prospective generalization.

\subsection{Compression Without Inference Loss}
\label{sec:compression_without_loss}

\begin{figure*}[!t]
    \centering
    \includegraphics[
        width=0.96\textwidth
    ]{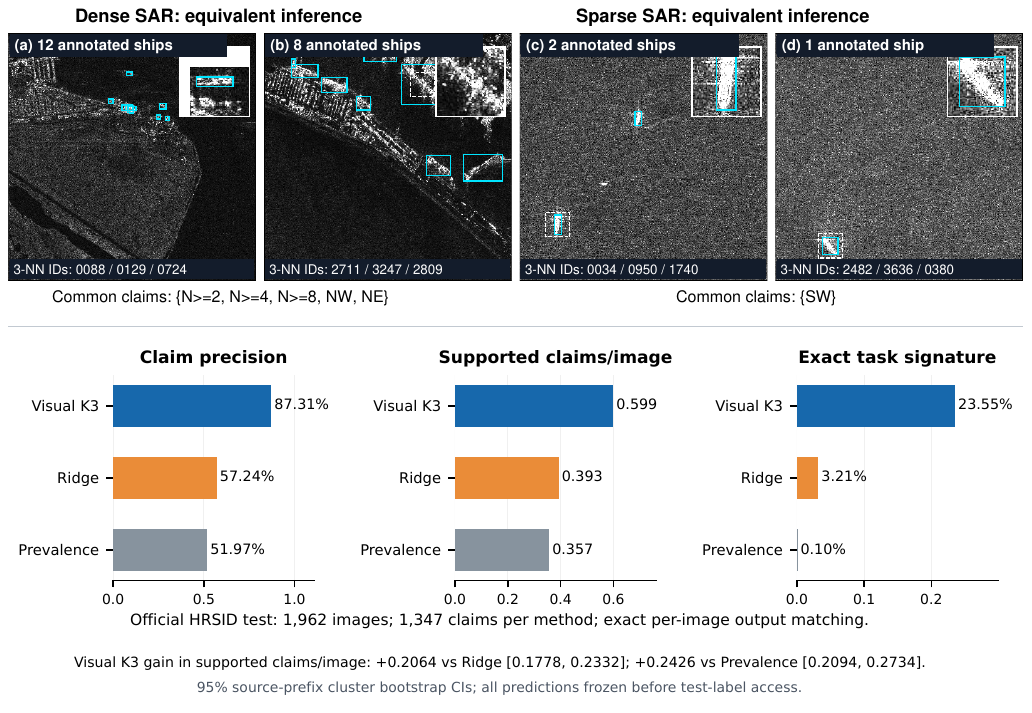}
    \caption{
        Inference equivalence and real-SAR inference
        on the official HRSID test split.
        Top: Dense and sparse SAR image pairs have
        distinct three-nearest-neighbor candidate
        sets but identical inference frontiers
        under seven registered task claims.
        Cyan boxes denote ground-truth annotations;
        examples were selected after evaluation.
        Bottom: Claim precision, supported claims
        per image, and exact task-signature recovery
        for three frozen methods on 1,962 test images,
        with 1,347 claims issued per method.
        Confidence intervals use source-prefix
        cluster bootstrap.
        Method differences reflect candidate-set
        construction, not gains from quotient coding.
    }
    \label{fig:real_sar_quotient}
    \label{fig:official_test}
\end{figure*}

We evaluate exact quotient coding across
the four registered observation regimes
of Sec.~\ref{sec:observation_ambiguity}.
Let $\mathfrak G_{\mathcal B}^{\mathrm{obs}}$
denote the distinct nonempty compatibility
sets observed in each regime.
The quotient code
$C_{\tau,\mathcal B}(\Gamma)=F_{\tau,\mathcal B}(\Gamma)$
assigns the same code to two compatibility sets
if and only if they support identical valid task claims.
We measure the reduction in required
distinct code values by
\begin{equation}
\label{eq:exp_quotient_reduction}
\begin{aligned}
N_\Gamma
&=|\mathfrak G_{\mathcal B}^{\mathrm{obs}}|,
\qquad
N_C
=|\mathfrak G_{\mathcal B}^{\mathrm{obs}}
/\!\sim_{\tau,\mathcal B}|,\\
R_{\mathrm{quot}}
&=1-\frac{N_C}{N_\Gamma}.
\end{aligned}
\end{equation}

Historically, 145 compatibility sets
yield 31 quotient codes (78.62\% fewer
required code values), with 17 multi-set
classes and 1,547 of 10,440 distinct-set
pairs sharing a code. Direct inference,
quotient decoding, and independent frontier
reconstruction agree on all 8,800
observations (zero mismatches), verifying
$D_\tau(C_{\tau,\mathcal B}(\Gamma))
=V_{\tau,\mathcal B}(\Gamma)$.
By minimality (Sec.~\ref{sec:quotient_coding}),
31 is the minimum number of required
code values for the observed historical
family, without claiming optimal storage
or computational complexity. OOD results and controls are reported in
Table~\ref{tab:exp_quotient}
(Appendix~\ref{app:quotient_statistics}).

The best in-sample cardinality-only decoder achieves
47.85\% exact-signature accuracy on historical observations.
Task closure strictly enlarges 129 of 145 observed compatibility
sets while preserving every valid task claim. Of 1,036 single-state deletions, 941 preserve exact inference,
whereas the other 95 make at least one claim unsupported
by the original compatibility set appear valid. Extending the
geometry task from nine to 13 claims refines
the historical quotient from 13 to 31 codes.
Further controls and witnesses are provided
in Appendices~\ref{app:cardinality_control}--\ref{app:task_refinement}.

On 1,962 official HRSID test images,
1,913 distinct 3-NN candidate
sets yield only 36 inference signatures
for a separate seven-claim SAR task.
The post hoc dense/sparse witnesses in
Fig.~\ref{fig:real_sar_quotient}
illustrate this bank-relative
inference equivalence, without implying
physical-state reconstruction.


\subsection{Accuracy and Specificity}
\label{sec:accuracy_specificity}

We evaluate four frozen compatibility-set predictors
trained under symmetric, coverage-weighted, closure-target,
and corridor-asymmetric objectives on 3,040 prospective
observations spanning geometry, mechanism, and joint-OOD,
using 13 registered task claims.

For each $o_i$, let $\Gamma_i$ and
$\widehat\Gamma_i$ denote the oracle and
predicted compatibility sets, with
$V_i=V_{\tau,\mathcal B}(\Gamma_i)$.
We set
$\widehat V_i=V_{\tau,\mathcal B}(\widehat\Gamma_i)$
for nonempty predictions and
$\widehat V_i=\varnothing$ otherwise
(abstention).
No fitting or threshold selection uses
the prospective tests; further prospective
evaluation details are provided in
Appendix~\ref{app:prospective_learning}.

We report exact set recovery
$P_\Gamma=\Pr[\widehat\Gamma_i=\Gamma_i]$,
coverage
$\mathrm{Cov}=\Pr[\Gamma_i\subseteq
\widehat\Gamma_i]$, and exact inference
recovery
$P_{\mathrm{exact}}=\Pr[\widehat V_i=V_i]$.
Since exact set recovery implies
exact inference recovery,
\begin{equation}
\begin{aligned}
\Delta_{\mathrm{infer}}
&=P_{\mathrm{exact}}-P_\Gamma\\
&=\Pr[\widehat{\Gamma}_i\neq\Gamma_i,\,
\widehat V_i=V_i].
\end{aligned}
\label{eq:exp43_inference_gap}
\end{equation}
Across the three OOD regimes, corridor-asymmetric learning
yields 89 cases with $\widehat{\Gamma}_i\neq\Gamma_i$ but
$\widehat V_i=V_i$, including 14 with at least one
nontrivial valid claim.
Across the four learners, exact-inference
recovery ranges from 92.00--96.00\% on
geometry-OOD, 93.88--94.50\% on
mechanism-OOD, and 92.81--93.75\% on
joint-OOD (Table~\ref{tab:exp43_learning}
in Appendix~\ref{app:prospective_learning}).

Falsifier incidence characterizes exact
inference preservation. For
$q\in\mathcal Q_\tau$, let
$Z_q=\mathcal B\setminus E_q$
and $\chi_{iq}=\mathbf{1}[
\Gamma_i\cap Z_q\neq\varnothing]$;
define $\widehat\chi_{iq}$ using
$\widehat\Gamma_i$. For nonempty $\widehat{\Gamma}_i$, we have
$q\in V_i\iff\chi_{iq}=0$ and
$q\in\widehat V_i\iff\widehat\chi_{iq}=0$;
hence
\begin{equation}
\widehat V_i=V_i
\iff
\widehat\chi_{iq}=\chi_{iq},
\quad\forall q\in\mathcal Q_\tau.
\label{eq:exp43_falsifier_equivalence}
\end{equation}
A registered-bank geometry-OOD witness
of exact inference despite incorrect
compatibility-set recovery is provided
in Appendix~\ref{app:inference_witness}.

For $N$ observations and
$Q=|\mathcal Q_\tau|$ claims, let
$a_i=\mathbf 1\{
\widehat\Gamma_i\ne\varnothing\}$.
Accounting for abstention, the normalized
unsupported-claim and lost-valid-claim
risks are
\begin{equation}
\begin{aligned}
R_{\mathrm{unsup}}
&=\frac{1}{NQ}\sum_{i,q}
a_i\chi_{iq}(1-\widehat\chi_{iq}),\\
R_{\mathrm{lost}}
&=\frac{1}{NQ}\sum_{i,q}
(1-\chi_{iq})
[1-a_i(1-\widehat\chi_{iq})].
\end{aligned}
\end{equation}
For nonempty predictions, missing
compatible falsifiers yield unsupported
claims, whereas spurious falsifiers
lose valid claims. Abstention affects
only lost-valid-claim risk, and
$R_{\mathrm{unsup}}\leq 1-\mathrm{Cov}$.

Closure-target attains the lowest
geometry-OOD unsupported risk (0.00115)
but the highest lost-valid risk (0.012404).
Corridor-asymmetric learning is
nondominated on geometry- and
mechanism-OOD, but not joint-OOD.
These findings are bank-relative;
specificity controls and complete risk
comparisons appear in
Appendix~\ref{app:specificity}.

On 1,962 official HRSID test images,
visual 3-NN achieves 87.31\% claim
precision, 0.599 supported claims/image,
and 23.55\% exact-signature recovery
at 1,347 issued claims/method
(Fig.~\ref{fig:real_sar_quotient}).
Supported-claim gains over ridge and
prevalence are 0.2064 (95\% CI:
0.1778--0.2332) and 0.2426
(0.2094--0.2734), respectively. These
reflect frozen candidate-set construction,
not quotient-coding gains.


\section{Conclusion}

Reliable inference need not reconstruct physical states.
We characterize exact valid-claim frontiers and the
coarsest deterministic code preserving them. Finite-bank results show fewer distinct code values and
exact inference despite set errors. Falsifier incidence
separates unsupported and lost claims; full-model
validity requires certified coverage.

\section*{LLM Usage Statement}
OpenAI's ChatGPT was used to edit or refine author-produced writing, related work and citations, code and implementation, and experimental design, execution, and results analysis. All scientific claims, theoretical statements, experimental designs, implementations, numerical results, and final manuscript content were reviewed and verified by the authors. The authors take full responsibility for the content of this paper.

\bibliographystyle{ieeenat_fullname}
\bibliography{main}


\clearpage
\appendix

\numberwithin{equation}{section}
\numberwithin{figure}{section}

\paragraph{Scope of theoretical and experimental guarantees.}
We distinguish the full physical state space
$\mathcal{S}_{\mathrm{phys}}$ from a finite registered
physical-state bank $\mathcal{B}$. The theoretical
results apply to the full physical space under their
stated assumptions, whereas the experimental
exactness claims are relative to the registered bank,
its observation mechanisms, and the specified task
claim family. In particular, exact validity or quotient
preservation on $\mathcal{B}$ does not imply the same
guarantee over unregistered physical states.
Extending finite-bank validity to the full physical
model requires an additional coverage or
falsification certificate.

\section{Additional Prospective Learning Results}
\label{app:prospective_learning}

\noindent\textit{Prospective evaluation protocol.}
We evaluate four frozen compatibility-set
learners: symmetric, coverage-weighted,
closure-target, and corridor-asymmetric.

The evaluation uses 3,040 prospective
observations across geometry-OOD
($N=1{,}600$), mechanism-OOD ($N=800$),
and joint-OOD ($N=640$), with
13 registered task claims.

For each observation $o_i$,
$\Gamma_i$ and $\widehat\Gamma_i$
denote the oracle and predicted
compatibility sets, respectively.
Their valid-inference sets are
\[
V_i=V_{\tau,\mathcal B}(\Gamma_i),
\qquad
\widehat V_i=
V_{\tau,\mathcal B}(\widehat\Gamma_i).
\]

All metrics are evaluated separately
within each OOD regime.
All predicted compatibility sets
are nonempty. Consequently, no
prospective sample invokes the
empty-set abstention rule, and
the reported claim risks coincide
with the nonempty-set
falsifier-incidence calculations.
No model fitting or threshold selection
is performed on the prospective tests.
Table~\ref{tab:exp43_learning}
reports the prospective evaluation results.

\setcounter{table}{0}
\renewcommand{\thetable}{A\arabic{table}}

\begin{table*}[t]
\centering
\caption{
Prospective results for four frozen learners across three OOD regimes.
$P_{\Gamma}$/$P_{\mathrm{exact}}$ denote exact set/inference recovery;
$R_{\mathrm{unsup}}$/$R_{\mathrm{lost}}$ denote unsupported/lost valid-claim rates;
$S_{\mathrm{sup}}$ is the mean maximum specificity of correctly supported claims.
}
\label{tab:exp43_learning}
\small
\setlength{\tabcolsep}{5pt}
\begin{tabular}{lcccccc}
\toprule
Method
& Cov.
& $P_{\Gamma}$
& $P_{\mathrm{exact}}$
& $10^3R_{\mathrm{unsup}}$
& $10^3R_{\mathrm{lost}}$
& $S_{\mathrm{sup}}$\\
\midrule
\multicolumn{7}{l}{\emph{Geometry-OOD ($N=1{,}600$)}}\\
Symmetric
& 0.9675 & 0.9488 & 0.9600 & 3.173 & 1.827 & 0.4872\\
Coverage-weighted
& 0.9588 & 0.9375 & 0.9538 & 3.269 & 2.115 & 0.4869\\
Closure-target
& 0.9938 & 0.8988 & 0.9200 & 1.154 & 12.404 & 0.4666\\
Corridor-asymmetric
& 0.9744 & 0.9294 & 0.9500 & 2.308 & 5.192 & 0.4797\\
\addlinespace[2pt]
\midrule
\multicolumn{7}{l}{\emph{Mechanism-OOD ($N=800$)}}\\
Symmetric
& 0.9400 & 0.9100 & 0.9388 & 6.346 & 2.404 & 0.4451\\
Coverage-weighted
& 0.9138 & 0.8938 & 0.9400 & 8.269 & 0.962 & 0.4508\\
Closure-target
& 0.9525 & 0.9075 & 0.9425 & 6.250 & 1.923 & 0.4463\\
Corridor-asymmetric
& 0.9225 & 0.9013 & 0.9450 & 7.019 & 0.962 & 0.4508\\
\addlinespace[2pt]
\midrule
\multicolumn{7}{l}{\emph{Joint-OOD ($N=640$)}}\\
Symmetric
& 0.9406 & 0.9094 & 0.9344 & 4.688 & 3.606 & 0.5188\\
Coverage-weighted
& 0.9313 & 0.9156 & 0.9375 & 5.168 & 1.683 & 0.5203\\
Closure-target
& 0.9625 & 0.8969 & 0.9281 & 4.327 & 4.327 & 0.5156\\
Corridor-asymmetric
& 0.9203 & 0.8953 & 0.9281 & 5.409 & 3.606 & 0.5188\\
\bottomrule
\end{tabular}
\end{table*}

For corridor-asymmetric learning,
33 geometry-OOD, 35 mechanism-OOD,
and 21 joint-OOD observations
(89 total) satisfy
$\widehat\Gamma_i\neq\Gamma_i$
but $\widehat V_i=V_i$.
Among these, 14 retain at least one
nontrivial valid task claim.

\section{Cardinality-Only Decoding Control}
\label{app:cardinality_control}

We test whether $|\Gamma|$ alone preserves
the exact valid-inference signature.
The optimal in-sample accuracy of a
cardinality-only decoder is
\begin{equation}
\label{eq:exp_cardinality_control}
A_{\mathrm{size}}
=\frac{1}{N}\sum_k\max_v n(k,v),
\end{equation}
Here $n(k,v)$ counts observations with
$|\Gamma|=k$ and exact valid-inference
signature $v$. The optimal historical
cardinality-only accuracy is $47.85\%$.
Regime-wise cardinality-only accuracies
and quotient-code counts for all four
observation regimes are reported in
Table~\ref{tab:exp_quotient}.
Thus, compatibility-set cardinality
alone cannot determine
valid-inference content.

\begin{figure*}[t]
    \centering
    \includegraphics[width=0.90\textwidth]  {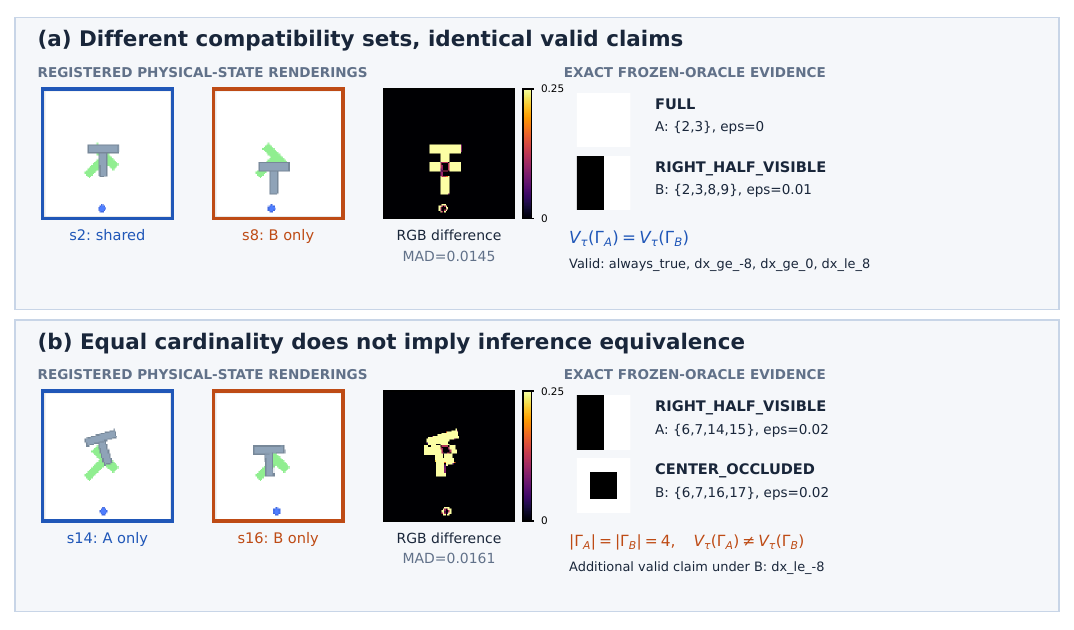}
    \caption{
    \textbf{Physical-state witnesses for inference equivalence.}
    (a) Distinct compatibility sets preserve identical
    valid claims; (b) equal-cardinality sets support
    different claims.
    Left: registered simulated RGB renderings,
    not oracle observation histories.
    Right: compatibility sets and valid claims
    independently evaluated by the frozen
    finite-bank oracle.
    }
    \label{fig:physical_inference_equivalence}
\end{figure*}

\section{Bank-Relative Task-Closure Analysis}
\label{app:task_closure}

The bank-relative task closure can enlarge
a compatibility set without changing
its valid-inference set:
\begin{equation}
\label{eq:exp_quotient_closure}
\begin{aligned}
\Gamma
&\subseteq K_{\tau,\mathcal B}(\Gamma)
=\!\!\bigcap_{q\in V_{\tau,\mathcal B}(\Gamma)}
\!\!\llbracket q\rrbracket_{\mathcal B},\\
V_{\tau,\mathcal B}(K_{\tau,\mathcal B}(\Gamma))
&=V_{\tau,\mathcal B}(\Gamma).
\end{aligned}
\end{equation}

Here, $\llbracket q\rrbracket_{\mathcal B}$
denotes the bank states satisfying $q$.
Every $q\in V_{\tau,\mathcal B}(\Gamma)$
holds on all states in $\Gamma$, hence
$\Gamma\subseteq K_{\tau,\mathcal B}(\Gamma)$.
Since validity is antitone under set
inclusion,
$V_{\tau,\mathcal B}(K_{\tau,\mathcal B}(\Gamma))
\subseteq V_{\tau,\mathcal B}(\Gamma)$.
By the definition of
$K_{\tau,\mathcal B}$, the reverse
inclusion also holds, proving equality.

The closure strictly enlarges 129 of
145 historical compatibility sets
($88.97\%$) without changing their
valid task claims. This demonstrates
that compatibility sets can contain
task-irrelevant physical distinctions.
Closure enlargement should not be
interpreted as compatibility-set
size compression.

\section{Single-State Deletion Analysis}
\label{app:state_deletion}

For $|\Gamma|>1$ and $s\in\Gamma$,
define the additional claims supported
after removing $s$:
\begin{equation}
\label{eq:exp_quotient_omission}
\mathcal U_{\tau,\mathcal B}(\Gamma,s)
=
V_{\tau,\mathcal B}(\Gamma\setminus\{s\})
\setminus V_{\tau,\mathcal B}(\Gamma).
\end{equation}

Since $\Gamma\setminus\{s\}\subseteq\Gamma$,
antitonicity of validity gives
$V_{\tau,\mathcal B}(\Gamma)
\subseteq
V_{\tau,\mathcal B}(\Gamma\setminus\{s\})$.
Consequently, deletion preserves the
exact valid-inference set iff
$\mathcal U_{\tau,\mathcal B}(\Gamma,s)
=\varnothing$.

Among 1,036 single-state deletions
from distinct nonsingleton historical
compatibility sets, 941 preserve
inference, whereas 95 introduce
previously unsupported claims.

In a two-state witness, removing one
compatible state newly asserts
\texttt{mode\_damped} without additional
measurements. Thus, quotient
identification must be distinguished
from arbitrary state deletion.

\section{Task Refinement and Quotient Splits}
\label{app:task_refinement}

We compare the nine-claim geometry task
$\tau_g$ with the full 13-claim task
$\tau_f$. Their quotient codes satisfy
the exact projection relation
\begin{equation}
\label{eq:exp_quotient_task_refinement}
C_{\tau_g,\mathcal B}
=
\pi_{\tau_g\leftarrow\tau_f}
\circ C_{\tau_f,\mathcal B}.
\end{equation}

For the same registered bank,
$\mathcal Q_{\tau_g}\subseteq
\mathcal Q_{\tau_f}$ gives
$V_{\tau_g,\mathcal B}(\Gamma)
=V_{\tau_f,\mathcal B}(\Gamma)
\cap\mathcal Q_{\tau_g}$.
Thus, full-task inference equivalence
implies geometry-task equivalence,
so each full-task quotient class has
a unique geometry-task image.
This defines
$\pi_{\tau_g\leftarrow\tau_f}$
and proves the projection relation.

Expanding the task increases the number
of distinct quotient code values from
$13$ to $31$ historically,
$10$ to $22$ for geometry-OOD,
$6$ to $15$ for mechanism-OOD, and
$8$ to $18$ for joint-OOD.

Among the geometry-task quotient
classes, 9 historical,
7 geometry-OOD, 5 mechanism-OOD,
and 6 joint-OOD classes each contain
multiple full-task quotient classes.

These finite-bank splits illustrate
the projection relation: compatibility
sets equivalent under the geometry task
can support different valid claims
under the expanded task.

\section{A Geometry-OOD Inference-Preservation Witness}
\label{app:inference_witness}

A geometry-OOD observation provides
a concrete example of exact inference
preservation without exact
compatibility-set recovery.

The oracle and predicted
compatibility sets satisfy
\begin{equation}
\begin{aligned}
\Gamma_i
&=\{3,4,5,7\}\\
&\subseteq
\widehat{\Gamma}_i
=\{2,3,4,5,6,7\}\\
&\subseteq
K_{\tau,\mathcal B}(\Gamma_i).
\end{aligned}
\end{equation}

By antitonicity and the task-closure
identity, these inclusions imply
identical valid-inference sets:
\begin{equation}
\widehat V_i=V_i
=
\{\texttt{always\_true},
  \texttt{dx\_le\_8}\}.
\end{equation}

The indices denote states in the
registered physical bank. The two
additional predicted candidates,
$2$ and $6$, do not falsify the
nontrivial valid claim
$\texttt{dx\_le\_8}$.

\section{Bank-Relative Specificity and Risk Diagnostics}
\label{app:specificity}

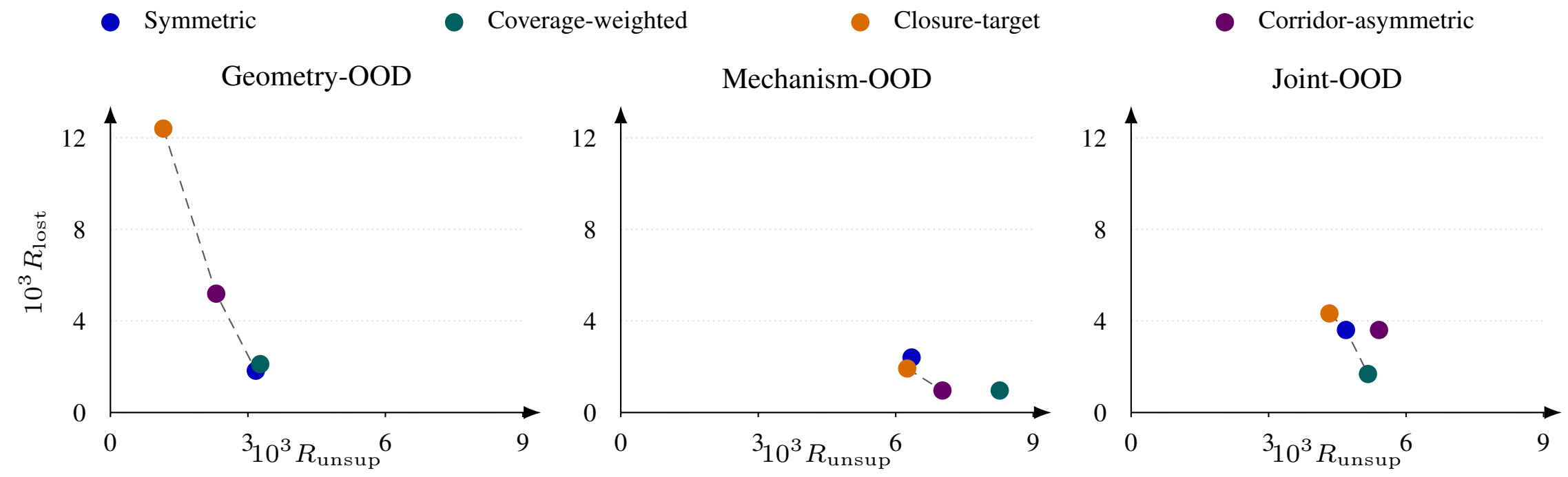
\begin{figure*}[t]
\centering
\begin{tikzpicture}[
    font=\scriptsize,
    >=Latex
]

\fill[blue!75!black]
    (0.8,4.45) circle (2.5pt);
\node[anchor=west] at (1.0,4.45)
    {Symmetric};

\fill[teal!75!black]
    (4.1,4.45) circle (2.5pt);
\node[anchor=west] at (4.3,4.45)
    {Coverage-weighted};

\fill[orange!85!black]
    (8.0,4.45) circle (2.5pt);
\node[anchor=west] at (8.2,4.45)
    {Closure-target};

\fill[violet!80!black]
    (11.5,4.45) circle (2.5pt);
\node[anchor=west] at (11.7,4.45)
    {Corridor-asymmetric};

\foreach \panel/\heading in {
    0.8/Geometry-OOD,
    5.7/Mechanism-OOD,
    10.6/Joint-OOD
}{
    \begin{scope}[
        shift={(\panel,0.70)},
        x=0.44cm,
        y=0.22cm
    ]

    \node[font=\footnotesize]
        at (4.5,14.6) {\heading};

    \foreach \yy in {0,4,8,12}{
        \draw[gray!25,densely dotted]
            (0,\yy) -- (9,\yy);
        \node[anchor=east]
            at (-0.25,\yy) {\yy};
    }

    \draw[->,thin]
        (0,0) -- (9.4,0);
    \draw[->,thin]
        (0,0) -- (0,13.4);

    \foreach \xx in {0,3,6,9}{
        \draw
            (\xx,0) -- (\xx,-0.20);
        \node[anchor=north]
            at (\xx,-0.35) {\xx};
    }

    \node at (4.5,-1.80)
        {$10^3R_{\mathrm{unsup}}$};

    \end{scope}
}

\node[rotate=90]
    at (0.05,2.15)
    {$10^3R_{\mathrm{lost}}$};

\begin{scope}[
    shift={(0.8,0.70)},
    x=0.44cm,
    y=0.22cm
]

\draw[black!65,densely dashed,thin]
    (1.154,12.404) --
    (2.308,5.192) --
    (3.173,1.827);

\fill[blue!75!black]
    (3.173,1.827) circle (2.5pt);
\fill[teal!75!black]
    (3.269,2.115) circle (2.5pt);
\fill[orange!85!black]
    (1.154,12.404) circle (2.5pt);
\fill[violet!80!black]
    (2.308,5.192) circle (2.5pt);

\end{scope}

\begin{scope}[
    shift={(5.7,0.70)},
    x=0.44cm,
    y=0.22cm
]

\draw[black!65,densely dashed,thin]
    (6.250,1.923) --
    (7.019,0.962);

\fill[blue!75!black]
    (6.346,2.404) circle (2.5pt);
\fill[teal!75!black]
    (8.269,0.962) circle (2.5pt);
\fill[orange!85!black]
    (6.250,1.923) circle (2.5pt);
\fill[violet!80!black]
    (7.019,0.962) circle (2.5pt);

\end{scope}

\begin{scope}[
    shift={(10.6,0.70)},
    x=0.44cm,
    y=0.22cm
]

\draw[black!65,densely dashed,thin]
    (4.327,4.327) --
    (4.688,3.606) --
    (5.168,1.683);

\fill[blue!75!black]
    (4.688,3.606) circle (2.5pt);
\fill[teal!75!black]
    (5.168,1.683) circle (2.5pt);
\fill[orange!85!black]
    (4.327,4.327) circle (2.5pt);
\fill[violet!80!black]
    (5.409,3.606) circle (2.5pt);

\end{scope}

\end{tikzpicture}

\caption{
Empirical unsupported-claim versus
lost-valid-claim risk across three prospective
OOD regimes.
Each panel compares four frozen
compatibility-set learners.
Both axes report risks multiplied by $10^3$,
with lower values preferred.
Dashed segments connect the observed
nondominated methods for visual guidance;
intermediate points are not asserted achievable.
Corridor-asymmetric learning is nondominated
under geometry- and mechanism-OOD, but is
dominated under joint-OOD.
These comparisons concern registered
finite-bank evaluations only.
}
\label{fig:ood_risk_pareto}
\end{figure*}

\begin{figure*}[!t]
    \centering
    \includegraphics[width=0.96\textwidth]
    {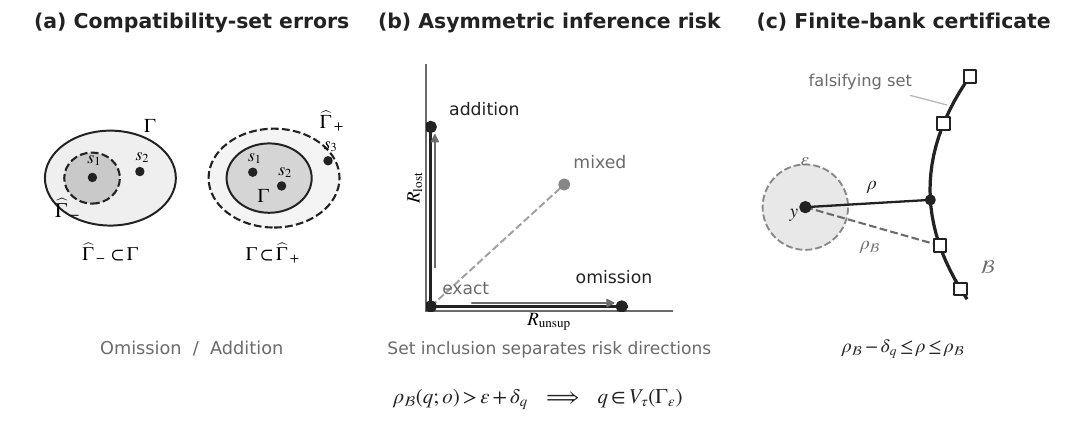}
    \caption{
        \textbf{Geometry of inference reliability.}
        (a) Omitting compatible states or
        adding incompatible states produces
        distinct set-inclusion errors.
        (b) These errors can respectively introduce
        unsupported claims or suppress valid claims,
        leading to asymmetric inference risks.
        (c) A finite-bank margin, together with a
        certified coverage bound, provides a
        sufficient condition for claim validity
        in the full physical model.
        All panels are schematic.
    }
    \label{fig:compatibility_errors}
\end{figure*}

For claim $q$, bank-relative
specificity is the fraction of
registered states falsifying $q$:
\begin{equation}
\operatorname{Spec}_{\mathcal B}(q)
=
1-\frac{
|\llbracket q\rrbracket_{\mathcal B}|
}{
|\mathcal B|
}.
\end{equation}

The supported specificity reported in
Table~\ref{tab:exp43_learning} is
\begin{equation}
S_{\mathrm{sup}}
=\frac{1}{N}\sum_{i=1}^{N}
\max_{q\in\widehat V_i\cap V_i}
\operatorname{Spec}_{\mathcal B}(q),
\label{eq:app_supported_specificity}
\end{equation}
where the maximum is defined as zero
when $\widehat V_i\cap V_i=\varnothing$.

We measure the mean maximum specificity
of unsupported predicted claims:
\begin{equation}
S_{\mathrm{unsupp}}
=
\frac{1}{N}
\sum_{i=1}^{N}
\max_{q\in\widehat V_i\setminus V_i}
\operatorname{Spec}_{\mathcal B}(q),
\end{equation}
Here too, an empty maximum is zero.

We also measure the fraction of
observations retaining at least one
correctly supported nontrivial claim:
\begin{equation}
P_{\mathrm{nt}}
=
\Pr\left[
\exists q\in\widehat V_i\cap V_i:
\operatorname{Spec}_{\mathcal B}(q)>0
\right].
\end{equation}

For corridor-asymmetric learning,
$P_{\mathrm{nt}}$ equals $0.7525$,
$0.8050$, and $0.8281$ across
geometry-, mechanism-, and joint-OOD,
respectively.

Under geometry-OOD,
\texttt{dx\_le\_8} and
\texttt{reference\_and\_dx\_ge\_0}
have equal specificity ($0.75$)
but unequal unsupported emission rates:
$0/294$ versus $4/140$.

Figure~\ref{fig:ood_risk_pareto}
shows that, with training fixed,
corridor-asymmetric learning is
nondominated in the empirical
$(R_{\mathrm{unsup}},R_{\mathrm{lost}})$
comparison under geometry- and
mechanism-OOD, but not joint-OOD.

These comparisons are finite-bank
and regime-specific.

\section{Temporal Observation Refinement Witness}
\label{app:temporal_refinement}

For observation $o=(y,M,h,\epsilon)$,
the bank falsification margin is
\begin{equation}
\rho_{\mathcal B}(q;o)
=
\min_{\substack{s\in\mathcal B\\q(s)=0}}
d_{M,h}\bigl(y,\mathcal O_h(s)\bigr),
\label{eq:app_bank_margin}
\end{equation}
where an empty minimum is $+\infty$.

Under checkerboard occlusion in an unseen
geometry, reference and damped dynamics
are indistinguishable in the first frame,
so $q=\texttt{mode\_reference}$ is
unsupported. The full pulse isolates one
compatible state:
\begin{equation}
\begin{aligned}
\rho_{\mathcal B}(q;o_{\mathrm{first}})
&=0\leq\epsilon,\\
\rho_{\mathcal B}(q;o_{\mathrm{full}})
&=0.0432>\epsilon=0.
\end{aligned}
\label{eq:app_physical_refinement}
\end{equation}

The refined frontier retains
$\texttt{dx\_ge\_16}$ and gains the
incomparable maximal claim
$\texttt{reference\_and\_dx\_ge\_0}$.

\FloatBarrier

\section{Mathematical Foundations and Proofs}
\label{app:mathematical_foundations}

\subsection{Physical Compatibility and Observation Consistency}
\label{app:physical_compatibility}

\noindent\textbf{Definition (Physical Compatibility).}
Let $\mathcal{S}_{\mathrm{phys}}$ denote the set of
physically admissible states, and let
$\mathcal{X}\subseteq\mathbb{R}^{d}$ be the observation
space. Throughout this appendix, we use
$H(m)$ and $O_\eta$ as shorthand for
the admissible acquisition-configuration family
$\mathcal H(m)$ and the physical observation
operator $\mathcal O_\eta$, respectively,
as defined in Sec.~3.1.
Thus,
$O_\eta:\mathcal S_{\mathrm{phys}}
\rightarrow\mathcal X$
for $\eta\in H(m)$.

For an observation $x$ and tolerance
$\epsilon\geq 0$, define the compatible-state set
\[
\Gamma_{\epsilon}(x;m)
=
\left\{
s\in\mathcal{S}_{\mathrm{phys}}:
\exists\eta\in H(m),\
\|O_{\eta}(s)-x\|\leq\epsilon
\right\}.
\]

\medskip
\noindent\textbf{Lemma (Generating-State Compatibility).}
Suppose the realized observation satisfies
\[
x=O_{\eta^\star}(s^\star)+e,
\]
where $s^\star\in\mathcal{S}_{\mathrm{phys}}$,
$\eta^\star\in H(m)$, and $\|e\|\leq\epsilon$.
Then
\[
s^\star\in\Gamma_{\epsilon}(x;m),
\qquad
\Gamma_{\epsilon}(x;m)\neq\varnothing.
\]

\noindent\textit{Proof.}
By the observation model,
\[
\|O_{\eta^\star}(s^\star)-x\|
=\|e\|\leq\epsilon.
\]
Since $\eta^\star\in H(m)$, the acquisition
configuration $\eta^\star$ witnesses the
existential condition defining
$\Gamma_{\epsilon}(x;m)$. Therefore,
$s^\star\in\Gamma_{\epsilon}(x;m)$, which
also establishes its nonemptiness.

\subsection{Factual Truth versus Inference Validity}
\label{app:factual_vs_validity}

\noindent\textbf{Definition (Inference Validity).}
Let $\mathcal{Q}_{\tau}$ be a finite family of
binary task claims
$q:\mathcal{S}_{\mathrm{phys}}\rightarrow\{0,1\}$.
For each claim $q$, define its truth set
\[
T_q=\{s\in\mathcal{S}_{\mathrm{phys}}:q(s)=1\}.
\]
For any nonempty compatibility set
$\Gamma\subseteq\mathcal{S}_{\mathrm{phys}}$,
the valid-inference set is
\[
V_{\tau}(\Gamma)
=
\{q\in\mathcal{Q}_{\tau}:\Gamma\subseteq T_q\}.
\]

\medskip
\noindent\textbf{Proposition (Validity Implies Factual Truth).}
Suppose the realized state satisfies
$s^\star\in\Gamma\neq\varnothing$.
Then, for every $q\in\mathcal{Q}_{\tau}$,
\[
q\in V_{\tau}(\Gamma)
\implies q(s^\star)=1.
\]
The converse does not hold in general.

\noindent\textit{Proof.}
By definition,
\[
q\in V_{\tau}(\Gamma)
\iff \Gamma\subseteq T_q.
\]
Since $s^\star\in\Gamma$, validity implies
$s^\star\in T_q$, and therefore
$q(s^\star)=1$.

To disprove the converse, consider two
physically admissible states $s_1\neq s_2$
and a common acquisition configuration $\eta_0$
such that
\[
O_{\eta_0}(s_1)=O_{\eta_0}(s_2)=x_0.
\]
Let $H(m)=\{\eta_0\}$, $\epsilon=0$, and
$\mathcal{S}_{\mathrm{phys}}=\{s_1,s_2\}$.
Then
\[
\Gamma_0(x_0;m)=\{s_1,s_2\}.
\]
Choose a task claim satisfying
\[
q(s_1)=1,\qquad q(s_2)=0.
\]
If the realized state is $s^\star=s_1$,
then $q(s^\star)=1$. However,
$s_2\in\Gamma_0(x_0;m)$ falsifies $q$, so
\[
q\notin V_{\tau}(\Gamma_0(x_0;m)).
\]
Thus factual truth does not imply
observation-supported validity.

\subsection{Physically Realizable Falsifying Measurements}
\label{app:physical_falsifiers}

\noindent\textbf{Definition (Physical Falsification Set).}
For a task claim $q\in\mathcal{Q}_{\tau}$ and
acquisition metadata $m$, define the admissible
falsifying state--configuration pairs by
\[
\mathcal{E}^{-}_q(m)
=
\left\{
(s,\eta)\in
\mathcal{S}_{\mathrm{phys}}\times H(m):
q(s)=0
\right\}.
\]
The corresponding physically realizable
falsifying measurement set is
\[
\mathcal{M}^{-}_q(m)
=
\left\{
O_{\eta}(s):
(s,\eta)\in\mathcal{E}^{-}_q(m)
\right\}.
\]

Define the falsification margin
\[
\rho(q;x,m)
=
\inf_{y\in\mathcal{M}^{-}_q(m)}
\|x-y\|,
\]
with $\inf\varnothing=+\infty$.

\medskip
\noindent\textbf{Lemma (Exact Falsifier Intersection).}
Assume that the compatibility set
$\Gamma_\epsilon(x;m)$ is nonempty.
Let $\overline{B}_{\epsilon}(x)$ denote the
closed observation ball
\[
\overline{B}_{\epsilon}(x)
=
\{y\in\mathcal{X}:\|x-y\|\leq\epsilon\}.
\]
For every $q\in\mathcal{Q}_{\tau}$,
\[
q\notin V_{\tau}(\Gamma_{\epsilon}(x;m))
\iff
\mathcal{M}^{-}_q(m)
\cap\overline{B}_{\epsilon}(x)
\neq\varnothing.
\]

\noindent\textit{Proof.}
By the definition of inference validity,
\[
\begin{aligned}
&q\notin V_{\tau}(\Gamma_{\epsilon}(x;m))\\
&\iff
\exists s\in\Gamma_{\epsilon}(x;m):
q(s)=0\\
&\iff
\exists(s,\eta)\in\mathcal{E}^{-}_q(m):
\|O_{\eta}(s)-x\|\leq\epsilon\\
&\iff
\mathcal{M}^{-}_q(m)
\cap\overline{B}_{\epsilon}(x)
\neq\varnothing.
\end{aligned}
\]
Each equivalence follows directly from the
definitions of compatibility and physically
realizable falsifying measurements.

\medskip
\noindent\textbf{Corollary (One-Sided Margin Conditions).}
For any physically realizable falsification set
and any observation--tolerance pair $(x,\epsilon)$
with $\epsilon\geq 0$ and
$\Gamma_\epsilon(x;m)\neq\varnothing$,
\[
\begin{aligned}
\rho(q;x,m)>\epsilon
&\implies
q\in V_{\tau}(\Gamma_{\epsilon}(x;m)),\\
\rho(q;x,m)<\epsilon
&\implies
q\notin V_{\tau}(\Gamma_{\epsilon}(x;m)).
\end{aligned}
\]

\noindent\textit{Proof.}
If $\rho(q;x,m)>\epsilon$, no falsifying
measurement lies within the closed observation
ball. The preceding lemma therefore implies
validity.

If $\rho(q;x,m)<\epsilon$, the definition of
the infimum guarantees a falsifying measurement
at distance strictly below $\epsilon$.
The preceding lemma then implies invalidity.

\subsection{Closedness and Exact Distance-Threshold Characterization}
\label{app:closed_falsification}

\noindent\textbf{Theorem (Exact Distance-Threshold Characterization).}
Fix acquisition metadata $m$, an observation
$x\in\mathcal{X}$, and tolerance $\epsilon\geq0$.
Assume $\Gamma_\epsilon(x;m)\neq\varnothing$.
Suppose the physically realizable falsification
set $\mathcal{M}^{-}_q(m)$ is closed in
$\mathbb{R}^{d}$. Then
\[
q\in V_{\tau}(\Gamma_{\epsilon}(x;m))
\iff
\rho(q;x,m)>\epsilon.
\]

\noindent\textit{Proof.}
If $\mathcal{M}^{-}_q(m)=\varnothing$,
no physically admissible state falsifies $q$.
Thus $q$ is valid, while
$\rho(q;x,m)=+\infty>\epsilon$.

Now suppose
$\mathcal{M}^{-}_q(m)\neq\varnothing$.
By the definition of the infimum, there
exists a sequence
$\{y_n\}\subseteq\mathcal{M}^{-}_q(m)$
such that
\[
\|x-y_n\|\longrightarrow\rho(q;x,m).
\]
The sequence is eventually bounded.
By the Bolzano--Weierstrass theorem, it
admits a convergent subsequence
$y_{n_k}\rightarrow y^\star$.
Since $\mathcal{M}^{-}_q(m)$ is closed,
\[
y^\star\in\mathcal{M}^{-}_q(m).
\]
Continuity of the norm gives
\[
\|x-y^\star\|=\rho(q;x,m).
\]
Hence the falsification distance is attained.

By the exact falsifier-intersection lemma,
\[
\begin{aligned}
&q\in V_{\tau}(\Gamma_{\epsilon}(x;m))\\
&\iff
\mathcal{M}^{-}_q(m)
\cap\overline{B}_{\epsilon}(x)
=\varnothing\\
&\iff
\min_{y\in\mathcal{M}^{-}_q(m)}
\|x-y\|>\epsilon\\
&\iff
\rho(q;x,m)>\epsilon.
\end{aligned}
\]
This proves the equivalence.

\medskip
\noindent\textbf{Counterexample (Failure without Closedness).}
Let
\[
\mathcal{S}_{\mathrm{phys}}
=
\{a\}\cup\{b_n:n\in\mathbb{N},\,n\geq1\},
\]
with a single acquisition configuration
and observation operator satisfying
\[
O(a)=0,\qquad O(b_n)=\frac{1}{n}.
\]
Consider the claim
\[
q(a)=1,\qquad q(b_n)=0
\quad\text{for all }n\geq1.
\]
Its falsification measurement set is
\[
\mathcal{M}^{-}_q
=
\left\{\frac{1}{n}:n\geq1\right\},
\]
which is not closed in $\mathbb{R}$.

At $x=0$ and $\epsilon=0$,
\[
\Gamma_0(0)=\{a\},
\qquad
q\in V_{\tau}(\Gamma_0(0)).
\]
Nevertheless,
\[
\rho(q;0)
=
\inf_{n\geq1}\frac{1}{n}
=0=\epsilon.
\]
Therefore
\[
q\in V_{\tau}(\Gamma_0(0)),
\qquad
\rho(q;0)\not>\epsilon.
\]
The exact distance-threshold equivalence
can thus fail without distance attainment.

\subsection{Sufficient Conditions for Closed Falsification Sets}
\label{app:falsifier_closedness_conditions}

\noindent\textbf{Proposition (Compact Falsifier Domain).}
Suppose the admissible falsifying
state--configuration set
$\mathcal{E}^{-}_q(m)$ is compact and
the joint observation map
\[
(s,\eta)\longmapsto O_{\eta}(s)
\]
is continuous on $\mathcal{E}^{-}_q(m)$.
Then the physically realizable falsifying
measurement set $\mathcal{M}^{-}_q(m)$
is compact and hence closed in
$\mathbb{R}^{d}$.

\noindent\textit{Proof.}
By definition,
\[
\mathcal{M}^{-}_q(m)
=
\left\{
O_{\eta}(s):
(s,\eta)\in\mathcal{E}^{-}_q(m)
\right\}.
\]
Thus $\mathcal{M}^{-}_q(m)$ is the
image of $\mathcal{E}^{-}_q(m)$
under the joint observation map.

Since $\mathcal{E}^{-}_q(m)$ is compact
and the observation map is continuous,
its image is compact.

Every compact subset of $\mathbb{R}^{d}$
is closed. Therefore
\[
\mathcal{M}^{-}_q(m)
\quad\text{is closed}.
\]

\medskip
\noindent\textbf{Corollary (Sufficient Primitive Conditions).}
Suppose:

\begin{enumerate}
\item
$\mathcal{S}_{\mathrm{phys}}$ is compact;

\item
$H(m)$ is compact;

\item
the falsifying-state set
$T_q^{c}=
\mathcal{S}_{\mathrm{phys}}\setminus T_q$
is closed relative to
$\mathcal{S}_{\mathrm{phys}}$;

\item
$(s,\eta)\mapsto O_{\eta}(s)$
is jointly continuous.
\end{enumerate}

Then $\mathcal{M}^{-}_q(m)$ is closed.

\noindent\textit{Proof.}
Because $T_q^{c}$ is closed in the
compact space $\mathcal{S}_{\mathrm{phys}}$,
it is compact.

Since $H(m)$ is compact,
\[
\mathcal{E}^{-}_q(m)
=
T_q^{c}\times H(m)
\]
is compact.

The preceding proposition therefore
implies that $\mathcal{M}^{-}_q(m)$
is compact and closed.

\medskip
\noindent\textbf{Counterexample (Compact State Space Is Insufficient).}
Let
\[
\mathcal{S}_{\mathrm{phys}}=[0,1],
\qquad H(m)=\{\eta_0\},
\]
and let the continuous observation
operator satisfy
\[
O_{\eta_0}(s)=s.
\]

Consider the binary task claim
\[
q(s)=
\begin{cases}
1, & s=0,\\
0, & s>0.
\end{cases}
\]

Although $\mathcal{S}_{\mathrm{phys}}$
and $H(m)$ are compact, the falsifying
state set is
\[
T_q^{c}=(0,1],
\]
which is not closed in
$\mathcal{S}_{\mathrm{phys}}$.

Consequently,
\[
\mathcal{M}^{-}_q(m)=(0,1]
\]
is not closed in $\mathbb{R}$.

Thus compactness of the full physical
state space and continuity of the
observation operator alone do not
guarantee closed falsification sets.

\subsection{Finite Registered Banks and Validity Certificates}
\label{app:finite_bank_validity}

\noindent\textbf{Definition (Bank-Restricted Compatibility).}
For fixed acquisition metadata $m$,
let
\[
B_m\subseteq
\mathcal{S}_{\mathrm{phys}}\times H(m)
\]
be a finite set of physically admissible
registered state--configuration pairs.

Define the bank-restricted compatibility set
\[
\Gamma_{\epsilon,B_m}(x;m)
=
\left\{
s:
\begin{array}{l}
\exists\eta,\ (s,\eta)\in B_m,\\
\|O_{\eta}(s)-x\|\leq\epsilon
\end{array}
\right\}.
\]

The corresponding bank-restricted
falsification measurement set is
\[
\mathcal{M}^{-}_{q,B_m}(m)
=
\left\{
O_{\eta}(s):
(s,\eta)\in B_m,\ q(s)=0
\right\}.
\]

Define its falsification margin by
\[
\rho_{B_m}(q;x,m)
=
\inf_{y\in\mathcal{M}^{-}_{q,B_m}(m)}
\|x-y\|,
\]
where $\inf\varnothing=+\infty$.

\medskip
\noindent\textbf{Proposition (Exact Bank Validity).}
For any observation $x$, tolerance $\epsilon\geq 0$,
and finite registered bank $B_m$ satisfying
$\Gamma_{\epsilon,B_m}(x;m)\neq\varnothing$,
\[
q\in V_{\tau}(\Gamma_{\epsilon,B_m}(x;m))
\iff
\rho_{B_m}(q;x,m)>\epsilon.
\]

\noindent\textit{Proof.}
Since $B_m$ is finite,
$\mathcal{M}^{-}_{q,B_m}(m)$
is a finite subset of $\mathbb{R}^{d}$.

Every finite subset of $\mathbb{R}^{d}$
is closed. Therefore the exact
distance-threshold characterization
proved above applies to
$\mathcal{M}^{-}_{q,B_m}(m)$, giving
\[
q\in V_{\tau}(\Gamma_{\epsilon,B_m}(x;m))
\iff
\rho_{B_m}(q;x,m)>\epsilon.
\]

\medskip
\noindent\textbf{Proposition (One-Sided Validity Inclusion).}
For every finite registered bank
$B_m\subseteq
\mathcal{S}_{\mathrm{phys}}\times H(m)$,
\[
\Gamma_{\epsilon,B_m}(x;m)
\subseteq
\Gamma_{\epsilon}(x;m).
\]
If $\Gamma_{\epsilon,B_m}(x;m)
\neq\varnothing$, then
\[
V_{\tau}(\Gamma_{\epsilon}(x;m))
\subseteq
V_{\tau}(\Gamma_{\epsilon,B_m}(x;m)).
\]

\noindent\textit{Proof.}
Every state--configuration pair in $B_m$
is physically admissible by assumption.
Hence any state compatible through a
registered pair is also compatible
in the full physical model:
\[
\Gamma_{\epsilon,B_m}(x;m)
\subseteq
\Gamma_{\epsilon}(x;m).
\]

Suppose
$q\in V_{\tau}(\Gamma_{\epsilon}(x;m))$.
Then
\[
\Gamma_{\epsilon}(x;m)\subseteq T_q.
\]
The compatibility-set inclusion implies
\[
\Gamma_{\epsilon,B_m}(x;m)\subseteq T_q,
\]
and therefore
\[
q\in V_{\tau}(\Gamma_{\epsilon,B_m}(x;m)).
\]
This establishes the claimed inclusion.

\medskip
\noindent\textbf{Corollary (Bank Falsification Certificate).}
If the bank falsification margin satisfies
\[
\rho_{B_m}(q;x,m)\leq\epsilon,
\]
then
\[
q\notin V_\tau(\Gamma_\epsilon(x;m)).
\]

\noindent\textbf{Proof.}
Since $B_m$ is finite,
the assumed margin bound implies that
there exists a registered pair
$(s,\eta)\in B_m$ satisfying
\[
q(s)=0,
\qquad
\|O_\eta(s)-x\|\leq\epsilon.
\]
Therefore,
\[
s\in\Gamma_{\epsilon,B_m}(x;m)
\subseteq\Gamma_\epsilon(x;m).
\]
Thus $s$ is a physically admissible
compatible state falsifying $q$, and hence
\[
q\notin V_\tau(\Gamma_\epsilon(x;m)).
\]

\subsection{Directed Falsifier Coverage and Full-Model Validity Certificates}
\label{app:directed_falsifier_coverage}

\noindent\textbf{Theorem (Directed Falsifier-Cover Bound).}
Let $\mathcal{M}^{-}_q(m)$ be nonempty,
and let $B_m$ be a finite registered bank.

Suppose there exists a certified constant
$\delta_q\in[0,+\infty)$ such that
\[
\forall y\in\mathcal{M}^{-}_q(m),
\quad
\exists b\in\mathcal{M}^{-}_{q,B_m}(m):
\|y-b\|\leq\delta_q.
\]

Then $\mathcal{M}^{-}_{q,B_m}(m)$
is nonempty, and for every observation $x$,
\[
\rho_{B_m}(q;x,m)-\delta_q
\leq
\rho(q;x,m)
\leq
\rho_{B_m}(q;x,m).
\]

\noindent\textit{Proof.}
Since $\mathcal{M}^{-}_q(m)$ is nonempty,
the assumed coverage condition guarantees
that $\mathcal{M}^{-}_{q,B_m}(m)$
is also nonempty.

Because every registered falsifying
measurement is physically admissible,
\[
\mathcal{M}^{-}_{q,B_m}(m)
\subseteq
\mathcal{M}^{-}_q(m).
\]
Taking infima over these sets gives
\[
\rho(q;x,m)
\leq
\rho_{B_m}(q;x,m).
\]

For the reverse bound, fix an arbitrary
$y\in\mathcal{M}^{-}_q(m)$.
By the coverage assumption, there exists
$b\in\mathcal{M}^{-}_{q,B_m}(m)$
such that
\[
\|y-b\|\leq\delta_q.
\]

By the triangle inequality,
\[
\begin{aligned}
\rho_{B_m}(q;x,m)
&\leq\|x-b\|\\
&\leq\|x-y\|+\|y-b\|\\
&\leq\|x-y\|+\delta_q.
\end{aligned}
\]

Since $y$ was arbitrary, taking the
infimum over all full-model falsifying
measurements yields
\[
\rho_{B_m}(q;x,m)
\leq
\rho(q;x,m)+\delta_q.
\]

Rearranging,
\[
\rho_{B_m}(q;x,m)-\delta_q
\leq
\rho(q;x,m).
\]

Combining both bounds establishes
\[
\rho_{B_m}-\delta_q
\leq\rho\leq\rho_{B_m}.
\]
This completes the proof.

\medskip
\noindent\textbf{Corollary (Full-Model Validity Certificate).}
Under the assumptions of the preceding theorem,
if $\Gamma_\epsilon(x;m)\neq\varnothing$ and
\[
\rho_{B_m}(q;x,m)
>
\epsilon+\delta_q,
\]
then
\[
q\in V_{\tau}(\Gamma_{\epsilon}(x;m)).
\]

\noindent\textit{Proof.}
The directed coverage bound implies
\[
\begin{aligned}
\rho(q;x,m)
&\geq\rho_{B_m}(q;x,m)-\delta_q\\
&>\epsilon.
\end{aligned}
\]
Therefore no physically realizable
falsifying measurement lies within
the closed observation ball of radius
$\epsilon$. By the exact
falsifier-intersection criterion,
$q$ is valid in the full physical model.

\medskip
\noindent\textbf{Remark (Empty Falsifier Sets).}
If $\mathcal{M}^{-}_q(m)=\varnothing$,
then the claim has no physically admissible
falsifier, and
\[
\rho(q;x,m)=\rho_{B_m}(q;x,m)=+\infty.
\]
The claim is globally valid without
requiring a covering bound.

Conversely, if
$\mathcal{M}^{-}_q(m)\neq\varnothing$
but
$\mathcal{M}^{-}_{q,B_m}(m)=\varnothing$,
no finite $\delta_q$ can satisfy the
directed coverage assumption.
In this case, bank validity alone does
not provide a full-model certificate.

\subsection{Finite-Bank Exactness versus Full-Model Exactness}
\label{app:bank_full_exactness}

\noindent\textbf{Proposition (Exact Bank--Full Inference Agreement).}
Fix an observation $(x,m)$ and
tolerance $\epsilon\geq0$.
Suppose
\[
\Gamma_{\epsilon,B_m}(x;m)\neq\varnothing.
\]

Then
\[
V_{\tau}(\Gamma_{\epsilon}(x;m))
=
V_{\tau}(\Gamma_{\epsilon,B_m}(x;m))
\]
if and only if every bank-valid claim
is also valid in the full physical model.

Equivalently, agreement holds if and only if
\[
\begin{aligned}
\forall q\in\mathcal{Q}_{\tau}:\quad
&q\in V_{\tau}(\Gamma_{\epsilon,B_m}(x;m))\\
&\implies
\mathcal{M}^{-}_q(m)
\cap\overline{B}_{\epsilon}(x)
=\varnothing.
\end{aligned}
\]

\noindent\textit{Proof.}
The previously established one-sided
validity inclusion gives
\[
V_{\tau}(\Gamma_{\epsilon}(x;m))
\subseteq
V_{\tau}(\Gamma_{\epsilon,B_m}(x;m)).
\]

Thus equality holds precisely when
the reverse inclusion also holds:
\[
V_{\tau}(\Gamma_{\epsilon,B_m}(x;m))
\subseteq
V_{\tau}(\Gamma_{\epsilon}(x;m)).
\]

This means that every bank-valid claim
must also be full-model valid.

By the exact falsifier-intersection
criterion, full-model validity is equivalent
to the absence of a physically realizable
falsifying measurement within the
closed observation ball.

This establishes both equivalences.

\medskip
\noindent\textbf{Corollary (Sufficient Coverage Certificate).}
Suppose that, for every bank-valid claim
$q\in\mathcal{Q}_{\tau}$, either

\begin{enumerate}
\item
$\mathcal{M}^{-}_q(m)=\varnothing$; or

\item
a finite directed falsifier-cover bound
$\delta_q$ has been certified and
\[
\rho_{B_m}(q;x,m)>\epsilon+\delta_q.
\]
\end{enumerate}

Then
\[
V_{\tau}(\Gamma_{\epsilon}(x;m))
=
V_{\tau}(\Gamma_{\epsilon,B_m}(x;m)).
\]

\noindent\textit{Proof.}
Consider any bank-valid claim $q$.

In the first case, no physically admissible
state falsifies $q$, so it is full-model valid.

In the second case, the directed
falsifier-cover theorem gives
\[
\rho(q;x,m)
\geq\rho_{B_m}(q;x,m)-\delta_q
>\epsilon.
\]
Hence $q$ is also full-model valid.

Therefore every bank-valid claim is
full-model valid. The preceding
proposition yields equality of the
two valid-inference sets.

\medskip
\noindent\textbf{Counterexample (Unregistered Physical Falsifier).}
Consider three physically admissible
states
\[
\mathcal{S}_{\mathrm{phys}}=\{a,b,c\}
\]
with one acquisition configuration and
observation operator
\[
O(a)=0,\qquad O(b)=0,\qquad O(c)=2.
\]

Let the task claim satisfy
\[
q(a)=1,\qquad q(b)=q(c)=0.
\]

Register only the states $a$ and $c$:
\[
B_m=\{(a,\eta_0),(c,\eta_0)\}.
\]

For $x=0$ and $\epsilon=0$,
\[
\Gamma_{\epsilon,B_m}(0;m)=\{a\},
\qquad
\Gamma_{\epsilon}(0;m)=\{a,b\}.
\]

The bank falsification margin is
\[
\rho_{B_m}(q;0,m)=2>\epsilon,
\]
so the claim is exactly valid on
the registered bank.

However, the unregistered state $b$
is compatible with the observation
and falsifies $q$. Consequently,
\[
\rho(q;0,m)=0,
\qquad
q\notin V_{\tau}(\Gamma_{\epsilon}(0;m)).
\]

Thus even a strictly positive
bank falsification margin does not,
by itself, certify full-model validity.

\subsection{Semantic Specificity Order and Downward-Closed Validity}
\label{app:semantic_specificity}

\noindent\textbf{Definition (Semantic Specificity).}
Let $\mathcal Q_\tau$ be the prescribed family
of binary task claims. For each
$q\in\mathcal Q_\tau$, define its truth set by
\[
T_q=
\{s\in\mathcal S_{\mathrm{phys}}:q(s)=1\}.
\]
For $q_1,q_2\in\mathcal Q_\tau$, define
\[
q_1\preceq_\tau q_2
\iff
T_{q_2}\subseteq T_{q_1}.
\]
Thus, $q_2$ is at least as semantically
specific as $q_1$ whenever the truth of
$q_2$ implies the truth of $q_1$.

Two claims are semantically identical if
\[
q_1\equiv_T q_2
\iff
T_{q_1}=T_{q_2}.
\]
Write
\[
\overline{\mathcal Q}_\tau
=
\mathcal Q_\tau/{\equiv_T}
\]
for the family of semantic equivalence
classes. The relation $\preceq_\tau$
induces an order on
$\overline{\mathcal Q}_\tau$.

\medskip
\noindent\textbf{Proposition (Semantic Partial Order).}
The relation $\preceq_\tau$ is a preorder
on $\mathcal Q_\tau$ and a partial order
on $\overline{\mathcal Q}_\tau$.

\noindent\textit{Proof.}
For every $q\in\mathcal Q_\tau$,
\[
T_q\subseteq T_q,
\]
so $q\preceq_\tau q$. Hence the
relation is reflexive.

If $q_1\preceq_\tau q_2$ and
$q_2\preceq_\tau q_3$, then
\[
T_{q_3}\subseteq T_{q_2}
\subseteq T_{q_1}.
\]
Therefore $q_1\preceq_\tau q_3$,
establishing transitivity.

If $q_1\preceq_\tau q_2$ and
$q_2\preceq_\tau q_1$, then
\[
T_{q_1}=T_{q_2}.
\]
This implies $q_1\equiv_T q_2$.
Consequently, antisymmetry holds on
the semantic equivalence classes,
and the induced relation on
$\overline{\mathcal Q}_\tau$
is a partial order.

\medskip
\noindent\textbf{Proposition (Downward-Closed Validity).}
For any nonempty compatibility set
$\Gamma\subseteq\mathcal S_{\mathrm{phys}}$,
the valid-inference family
\[
V_\tau(\Gamma)
=
\{q\in\mathcal Q_\tau:
\Gamma\subseteq T_q\}
\]
is downward closed under semantic
specificity. Specifically,
\[
q_2\in V_\tau(\Gamma),
\quad
q_1\preceq_\tau q_2
\quad\Longrightarrow\quad
q_1\in V_\tau(\Gamma).
\]

\noindent\textit{Proof.}
Suppose $q_2\in V_\tau(\Gamma)$.
By the definition of validity,
\[
\Gamma\subseteq T_{q_2}.
\]
Since $q_1\preceq_\tau q_2$,
the definition of semantic specificity
gives
\[
T_{q_2}\subseteq T_{q_1}.
\]
Combining these inclusions yields
\[
\Gamma\subseteq T_{q_2}
\subseteq T_{q_1}.
\]
Therefore $\Gamma\subseteq T_{q_1}$,
which is equivalent to
$q_1\in V_\tau(\Gamma)$.

\medskip
\noindent\textbf{Proposition (Antitonicity under
Compatibility Refinement).}
For any nonempty compatibility sets
$\Gamma_1,\Gamma_2$ satisfying
\[
\Gamma_2\subseteq\Gamma_1,
\]
the corresponding valid-inference
sets satisfy
\[
V_\tau(\Gamma_1)
\subseteq
V_\tau(\Gamma_2).
\]

\noindent\textit{Proof.}
Take any $q\in V_\tau(\Gamma_1)$.
By definition,
\[
\Gamma_1\subseteq T_q.
\]
Since $\Gamma_2\subseteq\Gamma_1$,
we obtain
\[
\Gamma_2\subseteq\Gamma_1
\subseteq T_q.
\]
Hence $q\in V_\tau(\Gamma_2)$.
Because $q$ was arbitrary,
\[
V_\tau(\Gamma_1)
\subseteq
V_\tau(\Gamma_2).
\]

\medskip
\noindent\textbf{Remark (Role of Finiteness).}
The preceding results do not require
$\mathcal Q_\tau$ to be finite.
Finiteness becomes essential for the
general existence and unique minimal
generating representation of the
Exact Inference Frontier established
in the following subsection.

\subsection{Exact Inference Frontier:
Existence and Minimal Generation}
\label{app:frontier_existence}

\noindent\textbf{Definition (Exact Inference Frontier).}
Assume that the prescribed task-claim
family $\mathcal Q_\tau$ is finite.

Let $\overline{\mathcal Q}_\tau$ be
the semantic partial order defined
in the preceding subsection. For a
nonempty compatibility set $\Gamma$,
write
\[
\overline V_\tau(\Gamma)
=
\{[q]_T:q\in V_\tau(\Gamma)\}.
\]

Define the Exact Inference Frontier as
the set of maximal valid semantic claims:
\[
F_\tau(\Gamma)
=
\operatorname{Max}_{\preceq_\tau}
\overline V_\tau(\Gamma).
\]

When the task claims have pairwise
distinct truth sets, the semantic
equivalence classes can be identified
with the original claims, recovering
the frontier definition in
Eq.~\eqref{eq:exact_inference_frontier}.

For any
$G\subseteq\overline{\mathcal Q}_\tau$,
define its downward closure by
\[
\downarrow G
=
\{p\in\overline{\mathcal Q}_\tau:
\exists g\in G,\ p\preceq_\tau g\}.
\]

\medskip
\noindent\textbf{Theorem (Existence and
Exact Generation).}
For every nonempty compatibility set
$\Gamma$, the Exact Inference Frontier
exists and satisfies
\[
\overline V_\tau(\Gamma)
=
\downarrow F_\tau(\Gamma).
\]

In particular, every valid semantic
claim is either a frontier element or
is semantically weaker than at least
one frontier element.

\noindent\textit{Proof.}
Because $\mathcal Q_\tau$ is finite,
the semantic quotient
$\overline{\mathcal Q}_\tau$ is finite.
Consequently,
$\overline V_\tau(\Gamma)$
is also finite.

First, consider the case
\[
\overline V_\tau(\Gamma)=\varnothing.
\]
Then its set of maximal elements
is empty, and
\[
F_\tau(\Gamma)=\varnothing,
\qquad
\downarrow F_\tau(\Gamma)=\varnothing.
\]
The claimed equality therefore holds.

Now suppose
$\overline V_\tau(\Gamma)\neq\varnothing$.
Choose any
$p\in\overline V_\tau(\Gamma)$.

If $p$ is maximal, then
$p\in F_\tau(\Gamma)$.

Otherwise, there exists
$p_1\in\overline V_\tau(\Gamma)$
such that
\[
p\prec_\tau p_1,
\]
where $\prec_\tau$ denotes the
strict part of the partial order.

If $p_1$ is not maximal, choose
$p_2\in\overline V_\tau(\Gamma)$
with
\[
p_1\prec_\tau p_2.
\]

Continue this construction whenever
the current element is not maximal.
Because the valid semantic-claim
set is finite, a strictly increasing
chain cannot continue indefinitely.

Hence the construction terminates
at some
$f\in F_\tau(\Gamma)$ satisfying
\[
p\preceq_\tau f.
\]
Since $p$ was arbitrary,
\[
\overline V_\tau(\Gamma)
\subseteq
\downarrow F_\tau(\Gamma).
\]

For the reverse inclusion, take
$p\in\downarrow F_\tau(\Gamma)$.
By definition, there exists
$f\in F_\tau(\Gamma)$ with
\[
p\preceq_\tau f.
\]
Every frontier element is valid.
The downward-closed validity
proposition therefore implies that
$p$ is also valid. Thus
\[
\downarrow F_\tau(\Gamma)
\subseteq
\overline V_\tau(\Gamma).
\]

Combining the two inclusions proves
\[
\overline V_\tau(\Gamma)
=
\downarrow F_\tau(\Gamma).
\]

\medskip
\noindent\textbf{Proposition (Frontier
Antichain Property).}
The Exact Inference Frontier is an
antichain in
$\overline{\mathcal Q}_\tau$.
That is, for any
$f_1,f_2\in F_\tau(\Gamma)$,
\[
f_1\preceq_\tau f_2
\quad\Longrightarrow\quad
f_1=f_2.
\]

\noindent\textit{Proof.}
Suppose $f_1,f_2$ belong to the
frontier and
$f_1\preceq_\tau f_2$.

If $f_1\neq f_2$, then
\[
f_1\prec_\tau f_2.
\]
Since $f_2$ is valid, this
contradicts the maximality of
$f_1$ in the valid semantic-claim set.

Therefore $f_1=f_2$, proving
that no two distinct frontier
elements are comparable.

\medskip
\noindent\textbf{Theorem (Unique Minimal
Generating Representation).}
Fix a nonempty compatibility set $\Gamma$
and assume that $\mathcal{Q}_\tau$ is finite.
Let $G$ be any subset of
$\overline{\mathcal Q}_\tau$ satisfying
\[
\downarrow G
=
\overline V_\tau(\Gamma).
\]
Then
\[
F_\tau(\Gamma)\subseteq G.
\]

Consequently, the Exact Inference
Frontier is the unique
inclusion-minimal generating set
of $\overline V_\tau(\Gamma)$.

Furthermore,
\[
|G|\geq |F_\tau(\Gamma)|,
\]
with equality if and only if
\[
G=F_\tau(\Gamma).
\]

\noindent\textit{Proof.}
Let $G$ satisfy
\[
\downarrow G
=
\overline V_\tau(\Gamma).
\]

Because every element of $G$
belongs to its own downward closure,
we have
\[
G\subseteq\downarrow G
=
\overline V_\tau(\Gamma).
\]
Thus every generator in $G$ is valid.

Now take an arbitrary
$f\in F_\tau(\Gamma)$.
Since the frontier is contained
in the valid-inference set,
\[
f\in\overline V_\tau(\Gamma)
=
\downarrow G.
\]

By the definition of downward closure,
there exists $g\in G$ such that
\[
f\preceq_\tau g.
\]

Both $f$ and $g$ are valid.
Since $f$ is maximal among all
valid semantic claims, it follows
that
\[
f=g.
\]

Therefore $f\in G$. Because
$f$ was arbitrary,
\[
F_\tau(\Gamma)\subseteq G.
\]

The preceding existence theorem
established that
$F_\tau(\Gamma)$ itself generates
the entire valid-inference set.
Hence it is contained in every
generating set and is itself
a generating set.

It follows that the frontier is
the unique inclusion-minimal
generating set.

Finally, because all sets under
consideration are finite,
\[
F_\tau(\Gamma)\subseteq G
\quad\Longrightarrow\quad
|F_\tau(\Gamma)|\leq |G|.
\]
Equality of cardinalities holds
precisely when $G=F_\tau(\Gamma)$.

\medskip
\noindent\textbf{Remark (Scope of Frontier Minimality).}
The preceding minimality concerns
generating sets of semantically
distinct task claims under the
specified partial order.

It does not establish minimal
binary description length,
minimum representation dimension,
or minimum computational cost.

Moreover, the uniqueness statement
is defined on semantic equivalence
classes. If syntactically different
claims share the same truth set,
their labels need not yield
a unique syntactic representation.

\subsection{Frontier Refinement and Equality}
\label{app:frontier_refinement}

\noindent\textbf{Theorem (Frontier Domination
Characterization).}
Assume that $\mathcal Q_\tau$ is finite.
For any two nonempty compatibility
sets $\Gamma_1,\Gamma_2$,
\[
\overline V_\tau(\Gamma_1)
\subseteq
\overline V_\tau(\Gamma_2)
\]
if and only if
\[
\forall f_1\in F_\tau(\Gamma_1),
\quad
\exists f_2\in F_\tau(\Gamma_2):
\quad
f_1\preceq_\tau f_2.
\]

\noindent\textit{Proof.}
First suppose
\[
\overline V_\tau(\Gamma_1)
\subseteq
\overline V_\tau(\Gamma_2).
\]

Take any
$f_1\in F_\tau(\Gamma_1)$.
Since every frontier element is valid,
\[
f_1\in\overline V_\tau(\Gamma_1)
\subseteq
\overline V_\tau(\Gamma_2).
\]

By the exact-generation theorem,
\[
\overline V_\tau(\Gamma_2)
=
\downarrow F_\tau(\Gamma_2).
\]
Therefore there exists
$f_2\in F_\tau(\Gamma_2)$
such that
\[
f_1\preceq_\tau f_2.
\]
This establishes the forward
implication.

Conversely, suppose that every
$f_1\in F_\tau(\Gamma_1)$
is dominated by some element
$f_2\in F_\tau(\Gamma_2)$.

Take any
$p\in\overline V_\tau(\Gamma_1)$.

By exact generation, there exists
$f_1\in F_\tau(\Gamma_1)$ satisfying
\[
p\preceq_\tau f_1.
\]

By assumption, there exists
$f_2\in F_\tau(\Gamma_2)$ satisfying
\[
f_1\preceq_\tau f_2.
\]

Transitivity gives
\[
p\preceq_\tau f_1
\preceq_\tau f_2.
\]

Since $f_2$ is valid under
$\Gamma_2$ and validity is
downward closed, $p$ is also
valid under $\Gamma_2$.

Thus
\[
\overline V_\tau(\Gamma_1)
\subseteq
\overline V_\tau(\Gamma_2).
\]

The argument also covers an empty
first frontier, for which the
condition is vacuous.

\medskip
\noindent\textbf{Corollary (Frontier under
Compatibility Refinement).}
If
\[
\varnothing\neq\Gamma_2\subseteq\Gamma_1,
\]
then
\[
\forall f_1\in F_\tau(\Gamma_1),
\quad
\exists f_2\in F_\tau(\Gamma_2):
\quad
f_1\preceq_\tau f_2.
\]

\noindent\textit{Proof.}
The antitonicity of valid inference
gives
\[
\overline V_\tau(\Gamma_1)
\subseteq
\overline V_\tau(\Gamma_2).
\]
The preceding theorem then
establishes the claimed
frontier domination.

\medskip
\noindent\textbf{Counterexample (Frontier
Inclusion Need Not Hold).}
Let
\[
\mathcal S_{\mathrm{phys}}
=
\{a,b,c\},
\]
and consider two task claims with
truth sets
\[
T_{q_{ab}}=\{a,b\},
\qquad
T_{q_a}=\{a\}.
\]

Their semantic specificity
relation satisfies
\[
q_{ab}\prec_\tau q_a.
\]

Consider the nested compatibility
sets
\[
\Gamma_1=\{a,b\},
\qquad
\Gamma_2=\{a\}.
\]

For $\Gamma_1$, only $q_{ab}$
is valid, so
\[
F_\tau(\Gamma_1)=\{q_{ab}\}.
\]

For $\Gamma_2$, both claims
are valid, but $q_a$ is
strictly more specific. Hence
\[
F_\tau(\Gamma_2)=\{q_a\}.
\]

Consequently,
\[
F_\tau(\Gamma_1)
\not\subseteq
F_\tau(\Gamma_2),
\]
although
\[
\overline V_\tau(\Gamma_1)
\subseteq
\overline V_\tau(\Gamma_2).
\]

Thus compatibility refinement
preserves valid-inference
inclusion, but need not preserve
frontier inclusion.

\medskip
\noindent\textbf{Theorem (Equality of
Exact Inference Frontiers).}
For any two nonempty compatibility
sets $\Gamma_1,\Gamma_2$,
\[
F_\tau(\Gamma_1)=F_\tau(\Gamma_2)
\]
if and only if
\[
\overline V_\tau(\Gamma_1)
=
\overline V_\tau(\Gamma_2).
\]

\noindent\textit{Proof.}
Suppose first that the frontiers
are equal. Exact generation gives
\[
\begin{aligned}
\overline V_\tau(\Gamma_1)
&=\downarrow F_\tau(\Gamma_1)\\
&=\downarrow F_\tau(\Gamma_2)\\
&=\overline V_\tau(\Gamma_2).
\end{aligned}
\]

Conversely, suppose that the
valid semantic-claim sets
are equal.

By definition, each frontier
is the set of maximal elements
of its valid semantic-claim set.
Equal subsets of the same
partial order have identical
maximal elements. Therefore
\[
F_\tau(\Gamma_1)=F_\tau(\Gamma_2).
\]

\medskip
\noindent\textbf{Example (Distinct Compatibility
Sets with an Identical Frontier).}
Let
\[
\mathcal S_{\mathrm{phys}}
=
\{a,b,c\},
\qquad
\mathcal Q_\tau=\{q_{ab}\},
\]
where
\[
T_{q_{ab}}=\{a,b\}.
\]

Consider two distinct
compatibility sets:
\[
\Gamma_1=\{a\},
\qquad
\Gamma_2=\{a,b\}.
\]

Both sets are nonempty, and
\[
\Gamma_1\neq\Gamma_2.
\]
Nevertheless,
\[
\Gamma_1\subseteq T_{q_{ab}},
\qquad
\Gamma_2\subseteq T_{q_{ab}}.
\]

Therefore
\[
V_\tau(\Gamma_1)
=
V_\tau(\Gamma_2)
=
\{q_{ab}\},
\]
and
\[
F_\tau(\Gamma_1)
=
F_\tau(\Gamma_2)
=
\{q_{ab}\}.
\]

The two compatibility sets
contain different physical
possibilities but support
exactly the same prescribed
task conclusion.

\subsection{Falsification-Margin
Lipschitzness and Frontier Stability}
\label{app:margin_stability}

\noindent\textbf{Proposition (Lipschitz
Continuity of Falsification Margins).}
Fix a task claim $q$ and acquisition
metadata $m$.

Suppose the physically realizable
falsifying-measurement set
$\mathcal M_q^-(m)$ is nonempty.

Then, for every $x,x'\in\mathbb R^d$,
\[
\left|
\rho(q;x,m)-\rho(q;x',m)
\right|
\leq
\|x-x'\|.
\]

Thus, the falsification margin is
1-Lipschitz in the observation.

\noindent\textit{Proof.}
For any
$y\in\mathcal M_q^-(m)$,
the triangle inequality gives
\[
\|x-y\|
\leq
\|x-x'\|+\|x'-y\|.
\]

Taking the infimum over all
$y\in\mathcal M_q^-(m)$ yields
\[
\rho(q;x,m)
\leq
\|x-x'\|+\rho(q;x',m).
\]

Interchanging $x$ and $x'$ gives
\[
\rho(q;x',m)
\leq
\|x-x'\|+\rho(q;x,m).
\]

Combining the two inequalities
establishes
\[
\left|
\rho(q;x,m)-\rho(q;x',m)
\right|
\leq
\|x-x'\|.
\]

The argument does not require
$\mathcal M_q^-(m)$ to be closed
or compact.

\medskip
\noindent\textbf{Remark (Empty
Falsification Sets).}
If
\[
\mathcal M_q^-(m)=\varnothing,
\]
then
\[
\rho(q;x,m)=+\infty
\]
for every observation $x$.

Such a claim has no physically
admissible falsifier and remains
valid for every compatible
observation.

The ordinary real-valued
Lipschitz inequality is not
applied to this case because
differences involving $+\infty$
are not defined.

\medskip
\noindent\textbf{Proposition (Joint
Observation--Tolerance Bound).}
If $\mathcal M_q^-(m)$ is nonempty,
then for any observations $x,x'$
and tolerances $\epsilon,\epsilon'
\geq 0$,
\[
\begin{aligned}
&
\left|
[\rho(q;x,m)-\epsilon]
-
[\rho(q;x',m)-\epsilon']
\right|\\
&\qquad\leq
\|x-x'\|+|\epsilon-\epsilon'|.
\end{aligned}
\]

\noindent\textit{Proof.}
By the triangle inequality,
\[
\begin{aligned}
&
\left|
[\rho(q;x,m)-\epsilon]
-
[\rho(q;x',m)-\epsilon']
\right|\\
&\leq
|\rho(q;x,m)-\rho(q;x',m)|
+
|\epsilon-\epsilon'|.
\end{aligned}
\]

Applying the preceding
1-Lipschitz bound gives
the claimed inequality.

\medskip
\noindent\textbf{Theorem (Local Stability
of Exact Inference).}
Assume that:

\begin{enumerate}
\item
$\mathcal Q_\tau$ is finite;

\item
for every $q\in\mathcal Q_\tau$,
the physically realizable
falsifying-measurement set
$\mathcal M_q^-(m)$ is closed;

\item
the reference observation
$(x_0,\epsilon_0)$ satisfies
$\epsilon_0\geq 0$ and
$\Gamma_{\epsilon_0}(x_0;m)
\neq\varnothing$.
\end{enumerate}

Define the reference
threshold-separation margin
\[
\alpha_\tau(x_0,\epsilon_0;m)
=
\min_{\substack{q\in\mathcal Q_\tau\\
\mathcal M_q^-(m)\neq\varnothing}}
\left|
\rho(q;x_0,m)-\epsilon_0
\right|,
\]
where the minimum over an empty
index set is defined as $+\infty$.

Suppose
\[
\alpha_\tau(x_0,\epsilon_0;m)>0.
\]

Then every observation--tolerance pair
$(x,\epsilon)$ satisfying
\[
\Gamma_\epsilon(x;m)\neq\varnothing,
\qquad
\epsilon\geq 0,
\]
and
\[
\|x-x_0\|+|\epsilon-\epsilon_0|
<
\alpha_\tau(x_0,\epsilon_0;m)
\]
has exactly the same valid task-claim set
as the reference:
\[
V_\tau(\Gamma_\epsilon(x;m))
=
V_\tau(\Gamma_{\epsilon_0}(x_0;m)).
\]

Moreover, the corresponding Exact
Inference Frontiers are identical:
\[
F_\tau(\Gamma_\epsilon(x;m))
=
F_\tau(\Gamma_{\epsilon_0}(x_0;m)).
\]

\noindent\textit{Proof.}
Consider an arbitrary task claim
$q\in\mathcal Q_\tau$.

If
$\mathcal M_q^-(m)=\varnothing$,
then $q$ is valid for every
observation-tolerance pair,
so its validity cannot change.

Now suppose
$\mathcal M_q^-(m)\neq\varnothing$.

By the definition of
$\alpha_\tau$,
\[
\left|
\rho(q;x_0,m)-\epsilon_0
\right|
\geq
\alpha_\tau(x_0,\epsilon_0;m)>0.
\]

The joint observation-tolerance
bound implies
\[
\begin{aligned}
&
\left|
[\rho(q;x,m)-\epsilon]
-
[\rho(q;x_0,m)-\epsilon_0]
\right|\\
&\leq
\|x-x_0\|+|\epsilon-\epsilon_0|\\
&<
\alpha_\tau(x_0,\epsilon_0;m).
\end{aligned}
\]

Consequently,
$\rho(q;x,m)-\epsilon$
and
$\rho(q;x_0,m)-\epsilon_0$
have the same strict sign.

Because $\mathcal M_q^-(m)$
is closed, the exact
distance-threshold criterion gives
\[
q\in V_\tau(\Gamma_\epsilon(x;m))
\iff
\rho(q;x,m)>\epsilon.
\]

The same criterion applies at
$(x_0,\epsilon_0)$.
Since the strict signs agree,
the validity of $q$ is unchanged.

As $q$ was arbitrary,
\[
V_\tau(\Gamma_\epsilon(x;m))
=
V_\tau(\Gamma_{\epsilon_0}(x_0;m)).
\]

Since both compatibility sets are nonempty, the equality-of-
frontiers theorem then yields
\[
F_\tau(\Gamma_\epsilon(x;m))
=
F_\tau(\Gamma_{\epsilon_0}(x_0;m)).
\]

\medskip
\noindent\textbf{Corollary (Locally
Nonempty Compatibility).}
Suppose the reference observation
has a strictly compatible
state-configuration witness:
\[
\exists(s_0,\eta_0),\quad
\eta_0\in H(m),
\quad
\|O_{\eta_0}(s_0)-x_0\|
<
\epsilon_0.
\]

Define
\[
\beta
=
\epsilon_0-
\|O_{\eta_0}(s_0)-x_0\|>0.
\]

Under the assumptions of the
preceding theorem, if
\[
\|x-x_0\|+|\epsilon-\epsilon_0|
<
\min\{\alpha_\tau,\beta\},
\]
then both compatibility
nonemptiness and the Exact
Inference Frontier are preserved.

\noindent\textit{Proof.}
The triangle inequality gives
\[
\begin{aligned}
\|O_{\eta_0}(s_0)-x\|
&\leq
\|O_{\eta_0}(s_0)-x_0\|
+\|x-x_0\|\\
&<
\epsilon_0-|\epsilon-\epsilon_0|\\
&\leq\epsilon.
\end{aligned}
\]

Hence
$s_0\in\Gamma_\epsilon(x;m)$,
so the perturbed compatibility
set is nonempty.

The preceding theorem then
establishes equality of the
Exact Inference Frontiers.

\medskip
\noindent\textbf{Counterexample
(Instability at the Threshold).}
Consider the finite physical
state space
\[
\mathcal S_{\mathrm{phys}}
=
\{a,b\},
\]
with one acquisition
configuration and observations
\[
O(a)=0,
\qquad
O(b)=1.
\]

Let the task family consist
of a single claim $q$ with
\[
q(a)=1,
\qquad
q(b)=0.
\]

Its falsifying-measurement set is
\[
\mathcal M_q^-=\{1\},
\]
which is closed.

Fix $\epsilon=1$ and
the reference observation $x_0=0$.
Then
\[
\rho(q;x_0)=1=\epsilon.
\]

Both physical states are
compatible with $x_0$, so
\[
q\notin V_\tau(\Gamma_1(0)).
\]

For any $0<\delta<1$,
consider the perturbed
observation
\[
x_\delta=-\delta.
\]

Now
\[
\rho(q;x_\delta)=1+\delta
>
\epsilon.
\]

The compatible-state set is
\[
\Gamma_1(x_\delta)=\{a\},
\]
and therefore
\[
q\in V_\tau(\Gamma_1(x_\delta)).
\]

Thus an arbitrarily small
observation perturbation changes
the valid-inference set and
its Exact Inference Frontier.

Closedness guarantees the exact
threshold criterion, but does
not guarantee stability when
a falsification margin lies
exactly on the threshold.

\subsection{Inference Equivalence and
Canonical Quotient Construction}
\label{app:inference_equivalence}

\noindent\textbf{Definition (Admissible
Compatibility-Set Family).}
Let
\[
\mathcal G
\subseteq
2^{\mathcal S_{\mathrm{phys}}}
\setminus\{\varnothing\}
\]
be a fixed family of nonempty
compatibility sets.

In Sec.~3.2,
$\mathcal G$ is induced by admissible
observation--metadata--tolerance triples
$(x,m,\epsilon)\in\mathcal A$.
The results below also apply to any
fixed family of nonempty state sets.

All quotient constructions are relative
to this domain and the prescribed
task family $\mathcal Q_\tau$.

\medskip
\noindent\textbf{Definition (Inference
Equivalence).}
For any
$\Gamma_1,\Gamma_2\in\mathcal G$,
define
\[
\Gamma_1\sim_\tau\Gamma_2
\iff
V_\tau(\Gamma_1)
=
V_\tau(\Gamma_2).
\]

Thus, two compatibility sets
are inference equivalent
precisely when they support
the same valid task claims.

\medskip
\noindent\textbf{Proposition (Inference
Equivalence Is an Equivalence Relation).}
The relation $\sim_\tau$
is reflexive, symmetric,
and transitive on $\mathcal G$.

\noindent\textit{Proof.}
Reflexivity follows from
$V_\tau(\Gamma)=V_\tau(\Gamma)$.
If $\Gamma_1\sim_\tau\Gamma_2$, then
$V_\tau(\Gamma_1)=V_\tau(\Gamma_2)$;
symmetry of equality gives
$\Gamma_2\sim_\tau\Gamma_1$.
Finally, if $\Gamma_1\sim_\tau\Gamma_2$
and $\Gamma_2\sim_\tau\Gamma_3$, then
$V_\tau(\Gamma_1)=V_\tau(\Gamma_2)
=V_\tau(\Gamma_3)$, implying
$\Gamma_1\sim_\tau\Gamma_3$.

Thus $\sim_\tau$ is
an equivalence relation.

\medskip
\noindent\textbf{Definition (Inference
Quotient Space).}
For each $\Gamma\in\mathcal G$,
define its inference-equivalence
class as
\[
[\Gamma]_{\sim_\tau}
=
\{\Gamma'\in\mathcal G:
\Gamma'\sim_\tau\Gamma\}.
\]

The corresponding quotient
space is
\[
\mathcal G/{\sim_\tau}
=
\{[\Gamma]_{\sim_\tau}:
\Gamma\in\mathcal G\}.
\]

Define the canonical quotient
map by
\[
\pi_\tau:\mathcal G
\longrightarrow
\mathcal G/{\sim_\tau},
\qquad
\pi_\tau(\Gamma)
=
[\Gamma]_{\sim_\tau}.
\]

\medskip
\noindent\textbf{Proposition (Canonical
Exact Inference Recovery).}
There exists a well-defined
deterministic decoder
\[
D_\tau:
\mathcal G/{\sim_\tau}
\longrightarrow
2^{\mathcal Q_\tau}
\]
such that, for every
$\Gamma\in\mathcal G$,
\[
D_\tau(\pi_\tau(\Gamma))
=
V_\tau(\Gamma).
\]

\noindent\textit{Proof.}
Define
$D_\tau([\Gamma]_{\sim_\tau})
=V_\tau(\Gamma)$.
If
$[\Gamma_1]_{\sim_\tau}
=[\Gamma_2]_{\sim_\tau}$,
then
$\Gamma_1\sim_\tau\Gamma_2$,
which implies
$V_\tau(\Gamma_1)=V_\tau(\Gamma_2)$.
Hence the decoder is independent
of the chosen representative
and is well defined.
For every $\Gamma\in\mathcal G$,
\[
D_\tau(\pi_\tau(\Gamma))
=D_\tau([\Gamma]_{\sim_\tau})
=V_\tau(\Gamma).
\]
Thus the canonical quotient
preserves every valid task claim
exactly.

\medskip
\noindent\textbf{Definition (Binary
Inference Signature).}
Suppose $\mathcal Q_\tau$
is finite.

For every $\Gamma\in\mathcal G$,
define its binary inference
signature by
\[
\sigma_\tau(\Gamma)
=
\bigl(
\mathbf 1\{\Gamma\subseteq T_q\}
\bigr)_{q\in\mathcal Q_\tau}.
\]

Each coordinate records
whether the corresponding
task claim is valid.

The associated signature
decoder is
\[
D_\tau^{\mathrm{sig}}(b)
=
\{q\in\mathcal Q_\tau:b_q=1\}.
\]

By construction,
\[
D_\tau^{\mathrm{sig}}
(\sigma_\tau(\Gamma))
=
V_\tau(\Gamma).
\]

\medskip
\noindent\textbf{Theorem (Equivalent
Inference Representations).}
For finite $\mathcal Q_\tau$
and any
$\Gamma_1,\Gamma_2\in\mathcal G$,
the following statements
are equivalent:

\begin{enumerate}
\item
$\Gamma_1\sim_\tau\Gamma_2$;

\item
$V_\tau(\Gamma_1)
=
V_\tau(\Gamma_2)$;

\item
$\sigma_\tau(\Gamma_1)
=
\sigma_\tau(\Gamma_2)$;

\item
$F_\tau(\Gamma_1)
=
F_\tau(\Gamma_2)$.
\end{enumerate}

\noindent\textit{Proof.}
The first two statements
are equivalent by definition.

For any task claim
$q\in\mathcal Q_\tau$,
the corresponding signature
coordinate is
\[
\sigma_\tau(\Gamma)_q
=
\mathbf 1\{q\in V_\tau(\Gamma)\}.
\]

Consequently, two signatures
are equal if and only if every
task claim has the same
validity under both
compatibility sets.

This is equivalent to equality
of their valid-inference sets,
proving equivalence of
statements 2 and 3.

The equality-of-frontiers
theorem established that
two compatibility sets have
the same Exact Inference
Frontier if and only if their
valid semantic-claim sets
are equal.

Because semantically identical
claims have identical truth
sets, their validity is
identical for every
compatibility set.

Thus equality of valid
semantic-claim classes is
equivalent to equality of
the original valid-claim sets.

Therefore statements 2
and 4 are equivalent.

Combining the implications
proves all four statements
are equivalent.

\medskip
\noindent\textbf{Theorem (Inference
Quotient--Signature Bijection).}
Suppose $\mathcal Q_\tau$ is finite.
The map
\[
\overline{\sigma}_\tau:
\mathcal G/{\sim_\tau}
\longrightarrow
\operatorname{im}\sigma_\tau,
\qquad
[\Gamma]_{\sim_\tau}
\longmapsto
\sigma_\tau(\Gamma)
\]
is well defined and bijective.

\noindent\textit{Proof.}
By the preceding theorem,
$\Gamma_1\sim_\tau\Gamma_2$
if and only if
$\sigma_\tau(\Gamma_1)
=\sigma_\tau(\Gamma_2)$.
Hence $\overline{\sigma}_\tau$
is well defined, and equality
of its outputs implies equality
of the input equivalence classes,
establishing injectivity.

For any
$b\in\operatorname{im}\sigma_\tau$,
there exists $\Gamma\in\mathcal G$
such that $b=\sigma_\tau(\Gamma)$.
Therefore,
\[
b=
\overline{\sigma}_\tau
([\Gamma]_{\sim_\tau}),
\]
establishing surjectivity.
Thus $\overline{\sigma}_\tau$
is a bijection.

\medskip
\noindent\textbf{Remark (Physical
and Finite-Bank Quotient Domains).}
The quotient construction is
relative to the chosen family
$\mathcal G$.

If $\mathcal G$ is generated
from the complete physical
state space, inference
equivalence concerns validity
in that full model.

If $\mathcal G$ instead
consists of compatibility
sets restricted to a registered
finite physical-state bank,
the resulting quotient is
exact only relative to
that registered bank.

The latter equivalence classes
must not be interpreted as
globally inference equivalent
without an additional
full-model validity argument.

\subsection{Coarsest Exact
Inference-Preserving Coding}
\label{app:coarsest_quotient}

\noindent\textbf{Definition (Deterministic
Exact Inference-Preserving Code).}
Let $\mathcal G$ be the fixed
family of admissible nonempty
compatibility sets.

A deterministic encoder
\[
r:\mathcal G\longrightarrow\mathcal Z
\]
is exact inference preserving
for $\mathcal Q_\tau$ if there
exists a deterministic decoder
\[
d:\operatorname{im}r
\longrightarrow 2^{\mathcal Q_\tau}
\]
such that
\[
d(r(\Gamma))=V_\tau(\Gamma)
\qquad
\forall\Gamma\in\mathcal G.
\]

No restriction is imposed on
the numerical form, dimension,
or computational implementation
of the code alphabet $\mathcal Z$.

\medskip
\noindent\textbf{Theorem (Necessary
Distinction between Inference Classes).}
Let $r$ be any deterministic
exact inference-preserving code.

Then, for all
$\Gamma_1,\Gamma_2\in\mathcal G$,
\[
r(\Gamma_1)=r(\Gamma_2)
\quad\Longrightarrow\quad
\Gamma_1\sim_\tau\Gamma_2.
\]

Equivalently,
\[
\Gamma_1\not\sim_\tau\Gamma_2
\quad\Longrightarrow\quad
r(\Gamma_1)\neq r(\Gamma_2).
\]

\noindent\textit{Proof.}
Since $r$ is exact inference
preserving, there exists a
deterministic decoder $d$
such that
$d(r(\Gamma))=V_\tau(\Gamma)$
for every $\Gamma\in\mathcal G$.
If $r(\Gamma_1)=r(\Gamma_2)$,
then
\[
\begin{aligned}
V_\tau(\Gamma_1)
&=d(r(\Gamma_1))\\
&=d(r(\Gamma_2))\\
&=V_\tau(\Gamma_2).
\end{aligned}
\]
Hence
$\Gamma_1\sim_\tau\Gamma_2$
by definition.
The second implication
follows by contraposition.

\medskip
\noindent\textbf{Theorem (Canonical
Factorization of Exact Codes).}
For every deterministic exact
inference-preserving code $r$,
there exists a unique
surjective map
\[
h:\operatorname{im}r
\longrightarrow
\mathcal G/{\sim_\tau}
\]
such that
\[
\pi_\tau=h\circ r.
\]

\noindent\textit{Proof.}
For any $z\in\operatorname{im}r$,
choose $\Gamma\in\mathcal G$
with $r(\Gamma)=z$ and define
$h(z)=[\Gamma]_{\sim_\tau}$.
If $r(\Gamma_1)=r(\Gamma_2)$,
the necessary-distinction theorem
implies $\Gamma_1\sim_\tau\Gamma_2$.
Hence $h$ is well defined.

For every $\Gamma\in\mathcal G$,
\[
h(r(\Gamma))
=[\Gamma]_{\sim_\tau}
=\pi_\tau(\Gamma),
\]
so $\pi_\tau=h\circ r$.
Every class $[\Gamma]_{\sim_\tau}$
is attained at $r(\Gamma)$;
therefore $h$ is surjective.

If another map $h'$ satisfies
$\pi_\tau=h'\circ r$, then for
any $z=r(\Gamma)\in\operatorname{im}r$,
\[
h'(z)
=h'(r(\Gamma))
=\pi_\tau(\Gamma)
=h(z).
\]
Thus $h'=h$, establishing
uniqueness.

\medskip
\noindent\textbf{Corollary (Unique
Coarsest Exact Partition).}
Let
\[
\mathcal P_r
=
\{r^{-1}(\{z\}):
z\in\operatorname{im}r\}
\]
be the partition induced by
an exact inference-preserving
encoder $r$.

Every cell of $\mathcal P_r$
is contained in one
inference-equivalence class.

Consequently, the quotient
partition
\[
\mathcal P_\tau
=
\mathcal G/{\sim_\tau}
\]
is the unique coarsest partition
among deterministic exact
inference-preserving partitions.

\noindent\textit{Proof.}
Let $r$ be any deterministic
exact inference-preserving code.
If $\Gamma_1,\Gamma_2$
belong to the same cell of
$\mathcal P_r$, then
$r(\Gamma_1)=r(\Gamma_2)$.
By the necessary-distinction
theorem,
$\Gamma_1\sim_\tau\Gamma_2$.
Thus every cell of $\mathcal P_r$
is contained in an
inference-equivalence class,
so $\mathcal P_r$ refines
$\mathcal P_\tau$.

The canonical quotient map
$\pi_\tau$ is itself exact
inference preserving, and its
induced partition is
$\mathcal P_\tau$.
Hence $\mathcal P_\tau$
is admissible and coarser
than every admissible exact
partition. By antisymmetry
of partition refinement,
it is the unique coarsest
such partition.

\medskip
\noindent\textbf{Theorem (Minimum
Number of Distinct Code Values).}
Assume $\mathcal Q_\tau$ is finite.

Let
\[
N_\tau^{\mathrm{code}}
=
|\mathcal G/{\sim_\tau}|
\]
be the number of distinct
inference-equivalence classes.

Then
\[
N_\tau^{\mathrm{code}}
=
\min_{\substack{
r:\mathcal G\to\mathcal Z\\
r\text{ exact inference preserving}
}}
|\operatorname{im}r|.
\]

The minimum is attained by
the canonical quotient map
$\pi_\tau$.

Moreover, let
$\operatorname{Idl}(\overline{\mathcal Q}_\tau)$
denote the family of downward-closed subsets
of the semantic claim poset. Then
\[
\begin{aligned}
N_\tau^{\mathrm{code}}
&\le
|\operatorname{Idl}(\overline{\mathcal Q}_\tau)|
\le 2^{|\overline{\mathcal Q}_\tau|}\\
&\le 2^{|\mathcal Q_\tau|}.
\end{aligned}
\]

\noindent\textit{Proof.}
Every valid semantic-claim set
$\overline V_\tau(\Gamma)$
is downward closed in
$\overline{\mathcal Q}_\tau$.
Moreover, the map
\[
[\Gamma]_{\sim_\tau}
\longmapsto
\overline V_\tau(\Gamma)
\]
is injective by the definition of inference
equivalence. Every ideal is a subset of
$\overline{\mathcal Q}_\tau$, and
$|\overline{\mathcal Q}_\tau|
\leq|\mathcal Q_\tau|$.
Thus the stated upper bound follows,
and $N_\tau^{\mathrm{code}}$ is finite.

Now let $r$ be any exact
inference-preserving encoder.

The canonical factorization
theorem provides a surjection
\[
h:\operatorname{im}r
\longrightarrow
\mathcal G/{\sim_\tau}.
\]

Since the quotient contains
$N_\tau^{\mathrm{code}}$ distinct classes,
the encoder must have at least
$N_\tau^{\mathrm{code}}$ distinct values:
\[
|\operatorname{im}r|
\ge N_\tau^{\mathrm{code}}.
\]

On the other hand, the
canonical quotient map
$\pi_\tau$ is exact inference
preserving and satisfies
\[
\operatorname{im}\pi_\tau
=
\mathcal G/{\sim_\tau}.
\]

Thus
\[
|\operatorname{im}\pi_\tau|
=
N_\tau^{\mathrm{code}}.
\]

The lower bound is therefore
attainable, proving
\[
\min_r|\operatorname{im}r|
=
N_\tau^{\mathrm{code}}.
\]

\medskip
\noindent\textbf{Theorem (Equivalence
of Minimum-Cardinality Codes).}
Assume $\mathcal Q_\tau$ is finite,
and let $r$ be an exact
inference-preserving encoder
satisfying
\[
|\operatorname{im}r|
=
N_\tau^{\mathrm{code}}.
\]

Then there exists a bijection
\[
b:
\mathcal G/{\sim_\tau}
\longrightarrow
\operatorname{im}r
\]
such that
\[
r=b\circ\pi_\tau.
\]

Thus every minimum-cardinality
exact code differs from the
canonical quotient code only
by a bijective relabeling
of its code values.

\noindent\textit{Proof.}
By canonical factorization,
there exists a surjection
\[
h:\operatorname{im}r
\longrightarrow
\mathcal G/{\sim_\tau}
\]
satisfying $\pi_\tau=h\circ r$.
Since $r$ has minimum
cardinality,
\[
|\operatorname{im}r|
=|\mathcal G/{\sim_\tau}|
=N_\tau^{\mathrm{code}}<\infty.
\]
Thus $h$ is bijective.
Setting $b=h^{-1}$ gives
\[
r=b\circ\pi_\tau,
\]
as required.

\medskip
\noindent\textbf{Remark (Scope of
Quotient Optimality).}
The preceding results establish:

\begin{enumerate}
\item
the coarsest deterministic
exact inference-preserving
partition;

\item
the minimum number of
distinct code values needed
for exact recovery of all
prescribed valid task claims.
\end{enumerate}

They do not establish
minimum binary code length,
minimum representation
dimension, minimum storage,
or minimum computational cost.

Those quantities require
additional coding,
representation, or
computational assumptions.

Furthermore, the value $N_\tau^{\mathrm{code}}$
depends on the chosen
compatibility-set domain $\mathcal G$.

A minimum established over
a finite collection of
registered compatibility
sets need not equal the
minimum over all physically
admissible compatibility sets.

\subsection{Task Expansion and
Refinement of Inference Quotients}
\label{app:task_expansion}

\noindent\textbf{Definition (Nested Task Families).}
Let
\[
\mathcal Q_0\subseteq\mathcal Q_1
\]
be two finite families of
binary task claims defined on
the same physical state space.

Fix the same admissible
compatibility-set family
$\mathcal G$ for both tasks.

For $i\in\{0,1\}$, write
\[
V_i(\Gamma)
=
\{q\in\mathcal Q_i:
\Gamma\subseteq T_q\}.
\]

Let $\sim_i$, $F_i$,
and $N_i^{\mathrm{code}}$ denote the
corresponding inference
equivalence, Exact Inference
Frontier, and number of
inference-equivalence classes.

Claims with identical truth
sets are identified when
forming the semantic
partial orders.

\medskip
\noindent\textbf{Proposition (Restriction
of Valid Inference).}
For every $\Gamma\in\mathcal G$,
\[
V_0(\Gamma)
=
V_1(\Gamma)\cap\mathcal Q_0.
\]

\noindent\textit{Proof.}
Take any $q\in\mathcal Q_0$.

Since
$\mathcal Q_0\subseteq\mathcal Q_1$,
the claim also belongs to
$\mathcal Q_1$.

Its validity depends only on
whether
\[
\Gamma\subseteq T_q,
\]
which is independent of
the other claims included
in the task family.

Therefore
\[
q\in V_0(\Gamma)
\iff
q\in V_1(\Gamma).
\]

Restricting the latter set
to $\mathcal Q_0$ proves
the claimed equality.

\medskip
\noindent\textbf{Theorem (Task Expansion
Refines Inference Equivalence).}
For any
$\Gamma_1,\Gamma_2\in\mathcal G$,
\[
\Gamma_1\sim_1\Gamma_2
\quad\Longrightarrow\quad
\Gamma_1\sim_0\Gamma_2.
\]

Consequently, every
equivalence class under
the expanded task family
is contained in one
equivalence class under
the original task family.

\noindent\textit{Proof.}
Suppose $\Gamma_1\sim_1\Gamma_2$.
Then $V_1(\Gamma_1)=V_1(\Gamma_2)$.
By the restriction proposition,
\[
\begin{aligned}
V_0(\Gamma_1)
&=V_1(\Gamma_1)\cap\mathcal Q_0\\
&=V_1(\Gamma_2)\cap\mathcal Q_0\\
&=V_0(\Gamma_2).
\end{aligned}
\]
Hence $\Gamma_1\sim_0\Gamma_2$.

\medskip
\noindent\textbf{Theorem (Natural
Projection between Task Quotients).}
The map
\[
P_{1\to0}:
\mathcal G/{\sim_1}
\longrightarrow
\mathcal G/{\sim_0}
\]
defined by
\[
P_{1\to0}([\Gamma]_{\sim_1})
=
[\Gamma]_{\sim_0}
\]
is well defined and surjective.

Consequently,
\[
N_0^{\mathrm{code}}
\leq N_1^{\mathrm{code}}.
\]

\noindent\textit{Proof.}
If
$[\Gamma_1]_{\sim_1}=[\Gamma_2]_{\sim_1}$,
then $\Gamma_1\sim_1\Gamma_2$.
By the task-refinement theorem,
$\Gamma_1\sim_0\Gamma_2$,
so
$[\Gamma_1]_{\sim_0}=[\Gamma_2]_{\sim_0}$.
Thus $P_{1\to0}$ is well defined.

For every
$[\Gamma]_{\sim_0}\in\mathcal G/{\sim_0}$,
the class $[\Gamma]_{\sim_1}$ maps to it,
establishing surjectivity.
Since both quotient spaces are finite,
$N_0^{\mathrm{code}}\le N_1^{\mathrm{code}}$.

\medskip
\noindent\textbf{Theorem (Criterion
for Strict Quotient Refinement).}
Under the preceding assumptions,
\[
N_1^{\mathrm{code}}
>
N_0^{\mathrm{code}}
\]
if and only if there exist
$\Gamma_a,\Gamma_b\in\mathcal G$
and a newly introduced claim
$q\in\mathcal Q_1\setminus
\mathcal Q_0$ such that
\[
\Gamma_a\sim_0\Gamma_b
\]
but
\[
\mathbf 1\{\Gamma_a\subseteq T_q\}
\neq
\mathbf 1\{\Gamma_b\subseteq T_q\}.
\]

\noindent\textit{Proof.}
Suppose the stated pair
$\Gamma_a,\Gamma_b$ and claim $q$ exist.
Since $q$ has different validity on
$\Gamma_a$ and $\Gamma_b$,
we have $\Gamma_a\not\sim_1\Gamma_b$,
whereas $\Gamma_a\sim_0\Gamma_b$.
Thus $P_{1\to0}$ maps two distinct
expanded-task classes to the same
original-task class and is not injective.
As it is surjective and both quotient
spaces are finite,
$N_1^{\mathrm{code}}>N_0^{\mathrm{code}}$.

Conversely, suppose
$N_1^{\mathrm{code}}>N_0^{\mathrm{code}}$.
The surjection $P_{1\to0}$ cannot be
injective, so some
$\Gamma_a,\Gamma_b\in\mathcal G$
satisfy
$\Gamma_a\sim_0\Gamma_b$ but
$\Gamma_a\not\sim_1\Gamma_b$.
Hence a claim $q\in\mathcal Q_1$
has different validity on these sets.
Since
$V_0(\Gamma_a)=V_0(\Gamma_b)$,
such a claim cannot belong to
$\mathcal Q_0$.
Therefore
$q\in\mathcal Q_1\setminus\mathcal Q_0$,
as required.

\medskip
\noindent\textbf{Theorem (Projection
of Exact Inference Frontiers).}
Let
$\overline{\mathcal Q}_0$ and
$\overline{\mathcal Q}_1$
be the semantic partial
orders induced by the
nested task families.

Identify
$\overline{\mathcal Q}_0$
with its natural subset of
$\overline{\mathcal Q}_1$.

For
$G\subseteq\overline{\mathcal Q}_1$,
define
\[
\downarrow_1 G
=
\{p\in\overline{\mathcal Q}_1:
\exists g\in G,\,
p\preceq_1 g\}.
\]

Then, for every
$\Gamma\in\mathcal G$,
\[
F_0(\Gamma)
=
\operatorname{Max}_{\preceq_0}
\left(
\downarrow_1 F_1(\Gamma)
\cap
\overline{\mathcal Q}_0
\right).
\]

\noindent\textit{Proof.}
Exact frontier generation and validity
restriction give
\[
\begin{aligned}
\overline V_0(\Gamma)
&=\overline V_1(\Gamma)
  \cap\overline{\mathcal Q}_0\\
&=\bigl(\downarrow_1 F_1(\Gamma)\bigr)
  \cap\overline{\mathcal Q}_0.
\end{aligned}
\]
Taking $\operatorname{Max}_{\preceq_0}$
of both sides and using the definition
of $F_0(\Gamma)$ proves the claim.

\medskip
\noindent\textbf{Example (Task-Dependent
Quotient Splitting).}
Let
\[
\mathcal S_{\mathrm{phys}}
=
\{a,b,c\},
\]
and consider two claims
whose truth sets satisfy
\[
T_{q_{ab}}=\{a,b\},
\qquad
T_{q_a}=\{a\}.
\]

Define
\[
\mathcal Q_0=\{q_{ab}\},
\qquad
\mathcal Q_1=\{q_{ab},q_a\}.
\]

Consider the admissible
compatibility-set family
\[
\mathcal G
=
\bigl\{
\Gamma_a,\Gamma_{ab}
\bigr\},
\]
where
\[
\Gamma_a=\{a\},
\qquad
\Gamma_{ab}=\{a,b\}.
\]

Under the original task family,
\[
V_0(\Gamma_a)
=
V_0(\Gamma_{ab})
=
\{q_{ab}\}.
\]

Hence
\[
\Gamma_a\sim_0\Gamma_{ab},
\qquad
N_0^{\mathrm{code}}=1.
\]

Under the expanded task family,
\[
V_1(\Gamma_a)
=
\{q_{ab},q_a\},
\]
whereas
\[
V_1(\Gamma_{ab})
=
\{q_{ab}\}.
\]

Therefore
\[
\Gamma_a\not\sim_1\Gamma_{ab},
\qquad
N_1^{\mathrm{code}}=2.
\]

The added claim $q_a$
distinguishes two compatibility
sets that the original task
could not distinguish.

Moreover,
\[
F_0(\Gamma_a)=\{q_{ab}\},
\qquad
F_1(\Gamma_a)=\{q_a\}.
\]

Thus
\[
F_1(\Gamma_a)\cap\overline{\mathcal Q}_0
=
\varnothing
\neq
F_0(\Gamma_a).
\]

A direct intersection of
frontiers does not generally
recover the original frontier.
The downward-closure and
maximal-element projection
proved above is required.

\medskip
\noindent\textbf{Remark (Task Dependence
of Quotient Optimality).}
Increasing the prescribed
task family can split existing
inference-equivalence classes,
but cannot merge classes
that were already distinct.

The refinement is non-strict,
$N_1^{\mathrm{code}}=N_0^{\mathrm{code}}$,
if and only if every added claim
$q\in\mathcal Q_1\setminus\mathcal Q_0$
has a validity indicator
$\mathbf 1\{\Gamma\subseteq T_q\}$
that is constant on each original
$\sim_0$-equivalence class.

These conclusions require
the admissible compatibility-set
domain $\mathcal G$ to remain
fixed across the task comparison.

\subsection{Task Closure and
Inference-Preserving State Expansion}
\label{app:task_closure_supplementary_proofs}

\noindent\textit{Recall (Task Closure).}
For every nonempty
$\Gamma\subseteq S_{\mathrm{phys}}$,
the task closure of Sec.~3.2
can equivalently be expressed as
\[
\operatorname{Cl}_\tau(\Gamma)
=K_\tau(\Gamma)
=\bigcap_{q\in V_\tau(\Gamma)}T_q,
\]
where the intersection over an empty
family is defined as $S_{\mathrm{phys}}$.
Thus $\operatorname{Cl}_\tau(\Gamma)$
contains exactly the physical states
satisfying every claim valid under
$\Gamma$.

\noindent\textbf{Theorem (Closure Operator Properties).}
On the domain
$2^{\mathcal S_{\mathrm{phys}}}
\setminus\{\varnothing\}$,
the task-closure map satisfies:

\begin{enumerate}
\item Extensivity:
$\Gamma\subseteq\mathrm{Cl}_\tau(\Gamma)$.

\item
Monotonicity:
\[
\Gamma_1\subseteq\Gamma_2
\Longrightarrow
\mathrm{Cl}_\tau(\Gamma_1)
\subseteq
\mathrm{Cl}_\tau(\Gamma_2).
\]

\item Idempotence:
$\mathrm{Cl}_\tau(\mathrm{Cl}_\tau(\Gamma))
=\mathrm{Cl}_\tau(\Gamma)$.
\end{enumerate}

\noindent\textit{Proof.}
For every $q\in V_\tau(\Gamma)$,
validity gives $\Gamma\subseteq T_q$.
Taking the intersection yields
$\Gamma\subseteq\mathrm{Cl}_\tau(\Gamma)$.
The empty-family convention
also preserves this inclusion.

If $\Gamma_1\subseteq\Gamma_2$,
then
$V_\tau(\Gamma_2)\subseteq V_\tau(\Gamma_1)$.
Taking intersections therefore gives
$\mathrm{Cl}_\tau(\Gamma_1)
\subseteq\mathrm{Cl}_\tau(\Gamma_2)$.

For idempotence, extensivity implies
\[
V_\tau(\mathrm{Cl}_\tau(\Gamma))
\subseteq V_\tau(\Gamma).
\]
Conversely, every claim valid under
$\Gamma$ is valid throughout
$\mathrm{Cl}_\tau(\Gamma)$
by the definition of closure.
Hence
\[
V_\tau(\mathrm{Cl}_\tau(\Gamma))
=V_\tau(\Gamma).
\]
Applying the closure definition again
establishes
$\mathrm{Cl}_\tau(
\mathrm{Cl}_\tau(\Gamma))
=\mathrm{Cl}_\tau(\Gamma)$.

\medskip
\noindent\textbf{Corollary (Exact
Inference Preservation).}
For every nonempty
compatibility set $\Gamma$,
\[
V_\tau(\mathrm{Cl}_\tau(\Gamma))
=
V_\tau(\Gamma).
\]

If $\mathcal Q_\tau$ is finite,
the corresponding frontiers
also satisfy
\[
F_\tau(\mathrm{Cl}_\tau(\Gamma))
=
F_\tau(\Gamma).
\]

\noindent\textit{Proof.}
The equality of valid-inference
sets was established in the
preceding theorem.

The equality-of-frontiers
theorem then yields
the second assertion.

\medskip
\noindent\textbf{Theorem (Closure
Characterization of Inference Equivalence).}
For any nonempty
$\Gamma_1,\Gamma_2
\subseteq\mathcal S_{\mathrm{phys}}$,
\[
\mathrm{Cl}_\tau(\Gamma_1)
=
\mathrm{Cl}_\tau(\Gamma_2)
\]
if and only if
\[
V_\tau(\Gamma_1)
=
V_\tau(\Gamma_2).
\]

Consequently, the quotient
$\mathcal G/\sim_\tau$
is canonically bijective to
$\{\operatorname{Cl}_\tau(\Gamma):
\Gamma\in\mathcal G\}$,
via
$[\Gamma]_{\sim_\tau}
\mapsto\operatorname{Cl}_\tau(\Gamma)$.

\noindent\textit{Proof.}
If $V_\tau(\Gamma_1)=V_\tau(\Gamma_2)$,
their task closures coincide by definition.
Conversely, equality of task closures
implies equality of valid-inference sets
by exact inference preservation.
Thus, $\Gamma_1\sim_\tau\Gamma_2$
if and only if their task closures coincide.

Consequently, the map
$[\Gamma]_{\sim_\tau}\mapsto
\mathrm{Cl}_\tau(\Gamma)$
is well defined and injective.
Surjectivity onto the stated set
of task closures follows by construction.

\medskip
\noindent\textbf{Theorem (Maximal
Inference-Preserving State Expansion).}
Let $\Gamma$ be a nonempty
compatibility set.

For any set $E\subseteq\mathcal S_{\mathrm{phys}}$
satisfying $\Gamma\subseteq E$,
the following statements are equivalent:
 
\begin{enumerate}
\item
$V_\tau(E)=V_\tau(\Gamma)$;

\item
$E\subseteq
\mathrm{Cl}_\tau(\Gamma)$.
\end{enumerate}

Consequently,
$\mathrm{Cl}_\tau(\Gamma)$
is the unique largest
inference-preserving expansion
of $\Gamma$ among arbitrary
subsets of the physical
state space.

\noindent\textit{Proof.}
If $V_\tau(E)=V_\tau(\Gamma)$,
then $E\subseteq T_q$ for every
$q\in V_\tau(\Gamma)$.
Hence
$E\subseteq\mathrm{Cl}_\tau(\Gamma)$.

Conversely, suppose
$\Gamma\subseteq E\subseteq
\mathrm{Cl}_\tau(\Gamma)$.
Antitonicity gives
\[
\begin{aligned}
V_\tau(\Gamma)
&\supseteq V_\tau(E)\\
&\supseteq V_\tau(\mathrm{Cl}_\tau(\Gamma))
=V_\tau(\Gamma),
\end{aligned}
\]
where the equality follows from exact
inference preservation.
Thus $V_\tau(E)=V_\tau(\Gamma)$.

Finally, extensivity and exact inference
preservation make
$\mathrm{Cl}_\tau(\Gamma)$
an admissible inference-preserving expansion.
Every such expansion is contained in it,
so it is the unique largest expansion.

\medskip
\noindent\textbf{Counterexample (Task Closure
Need Not Be Observation Realizable).}
Consider
$\mathcal S_{\mathrm{phys}}=\{a,b,c\}$
with scalar observation operator
\[
O(a)=0,\qquad O(b)=2,\qquad O(c)=1.
\]
Let the task family contain a single claim
$q$ with truth set $T_q=\{a,b\}$.
For observation $x=0$ and tolerance
$\epsilon=0$, we have
$\Gamma_0(0)=\{a\}$ and
\[
\mathrm{Cl}_\tau(\Gamma_0(0))
=T_q=\{a,b\}.
\]

Suppose some observation $x'$ and
tolerance $\epsilon'\geq0$ realized
$\{a,b\}$ as a compatibility set.
Then the closed interval
$[x'-\epsilon',x'+\epsilon']$
would contain both $0$ and $2$.
By convexity, it would also contain
$1=O(c)$, making $c$ compatible,
a contradiction.
Hence this task closure is not
observation realizable.

\medskip
\noindent\textbf{Remark (Semantic Closure
versus Physical Reconstruction).}
Task closure is determined by prescribed task
truth sets, not by inversion of the physical
observation operator.

The maximality result concerns unrestricted
state-set expansions; the closure need not be
a physically realizable compatibility set.

Therefore exact inference
preservation does not imply
exact recovery of the
physical compatibility set.

\subsection{State Changes and Falsifier Incidence}
\label{app:state_transformations}

\noindent\textbf{Definition (Falsifying-State
Witness Set).}
For any nonempty state set
$\Gamma\subseteq\mathcal S_{\mathrm{phys}}$
and task claim $q\in\mathcal Q_\tau$,
define
\[
W_q(\Gamma)
=
\Gamma\cap T_q^c,
\]
where
\[
T_q^c
=
\mathcal S_{\mathrm{phys}}\setminus T_q.
\]

The set $W_q(\Gamma)$ contains
all states in $\Gamma$ that
falsify claim $q$.

By the definition of
inference validity,
\[
q\in V_\tau(\Gamma)
\iff
W_q(\Gamma)=\varnothing.
\]

\medskip
\noindent\textbf{Theorem (Exact
Falsifier-Incidence Criterion).}
For any two nonempty
state sets $\Gamma_1,\Gamma_2$,
\[
V_\tau(\Gamma_1)
=
V_\tau(\Gamma_2)
\]
if and only if, for every
$q\in\mathcal Q_\tau$,
\[
W_q(\Gamma_1)=\varnothing
\iff
W_q(\Gamma_2)=\varnothing.
\]

Equivalently,
\[
\mathbf 1
\{W_q(\Gamma_1)\neq\varnothing\}
=
\mathbf 1
\{W_q(\Gamma_2)\neq\varnothing\}
\]
for every prescribed claim.

\noindent\textit{Proof.}
For each $i\in\{1,2\}$ and
$q\in\mathcal Q_\tau$,
\[
q\in V_\tau(\Gamma_i)
\iff \Gamma_i\subseteq T_q
\iff W_q(\Gamma_i)=\varnothing.
\]
Thus equality of the valid-inference
sets is equivalent to matching
emptiness, or nonemptiness, indicators
for every prescribed claim.

\medskip
\noindent\textbf{Theorem (Exact Effect
of State Addition).}
Let $\Gamma$ be nonempty
and let
\[
A\subseteq
\mathcal S_{\mathrm{phys}}
\setminus\Gamma
\]
be a set of additional states.

Define
\[
\Gamma^+=\Gamma\cup A.
\]

Then
\[
V_\tau(\Gamma^+)
\subseteq V_\tau(\Gamma),
\]
and the valid claims lost
through state addition are
exactly
\[
\begin{aligned}
&V_\tau(\Gamma)
\setminus V_\tau(\Gamma^+)\\
&=
\{q\in V_\tau(\Gamma):
A\cap T_q^c\neq\varnothing\}.
\end{aligned}
\]

Consequently,
\[
V_\tau(\Gamma^+)=V_\tau(\Gamma)
\iff
A\subseteq\mathrm{Cl}_\tau(\Gamma).
\]

\noindent\textit{Proof.}
Since $\Gamma\subseteq\Gamma^+$,
antitonicity gives
$V_\tau(\Gamma^+)\subseteq V_\tau(\Gamma)$.
For any $q\in V_\tau(\Gamma)$,
we have $\Gamma\subseteq T_q$, hence
\[
\begin{aligned}
q\notin V_\tau(\Gamma^+)
&\iff A\not\subseteq T_q\\
&\iff A\cap T_q^c\neq\varnothing.
\end{aligned}
\]
This proves the exact lost-claim identity.

No valid claim is lost precisely when
$A\subseteq T_q$ for every
$q\in V_\tau(\Gamma)$, equivalently,
\[
A\subseteq
\bigcap_{q\in V_\tau(\Gamma)}T_q
=\mathrm{Cl}_\tau(\Gamma).
\]
Since state addition cannot create
new valid claims, this condition
is equivalent to
$V_\tau(\Gamma^+)=V_\tau(\Gamma)$.

\medskip
\noindent\textbf{Theorem (Exact Effect
of State Deletion).}
Let $\Gamma$ be nonempty,
and let $D\subseteq\Gamma$
be a deletion set such that
\[
\Gamma^-=\Gamma\setminus D
\neq\varnothing.
\]

Then
\[
V_\tau(\Gamma)
\subseteq V_\tau(\Gamma^-).
\]

The newly supported claims
are exactly
\[
\begin{aligned}
&V_\tau(\Gamma^-)
\setminus V_\tau(\Gamma)\\
&=
\{q\notin V_\tau(\Gamma):
W_q(\Gamma)\subseteq D\}.
\end{aligned}
\]

In particular,
\[
V_\tau(\Gamma^-)=V_\tau(\Gamma)
\]
if and only if
\[
\forall q\notin V_\tau(\Gamma),
\quad
W_q(\Gamma)\not\subseteq D.
\]

\noindent\textit{Proof.}
Since $\Gamma^-\subseteq\Gamma$,
antitonicity yields
$V_\tau(\Gamma)\subseteq V_\tau(\Gamma^-)$.
For each $q\in\mathcal Q_\tau$,
\[
W_q(\Gamma^-)
=(\Gamma\setminus D)\cap T_q^c
=W_q(\Gamma)\setminus D.
\]

For any $q\notin V_\tau(\Gamma)$,
\[
\begin{aligned}
q\in V_\tau(\Gamma^-)
&\iff W_q(\Gamma^-)=\varnothing\\
&\iff W_q(\Gamma)\subseteq D.
\end{aligned}
\]
This establishes the exact
newly supported-claim identity.
Since deletion cannot invalidate
previously valid claims, the
valid-inference sets coincide precisely
when no previously invalid claim
satisfies $W_q(\Gamma)\subseteq D$.

\medskip
\noindent\textbf{Corollary (Asymmetric
Effects of One-Sided Set Errors).}
For nonempty state sets:

\begin{enumerate}
\item
Adding states can destroy
valid claims but cannot
create newly valid claims.

\item
Deleting states can create
newly valid claims but cannot
destroy previously valid claims.

\item
Either operation can preserve
all valid task conclusions
without preserving the
original state set exactly.
\end{enumerate}

\noindent\textit{Proof.}
The first two statements follow
from the preceding addition and
deletion theorems.

For the third, consider the
three-state counterexample above.
The distinct sets $\{a\}$ and
$\{a,b\}$ support the same sole
task claim $q$.
Thus adding or removing $b$
changes the compatibility set
without changing its valid-inference set.

\noindent\textbf{Remark (Mixed State-Set
Errors).}
When states are both added and removed,
the one-sided monotonicity results
do not directly compare the initial
and final valid-inference sets.

Nevertheless, exact inference is
preserved if and only if
\[
W_q(\Gamma_1)=\varnothing
\iff W_q(\Gamma_2)=\varnothing
\]
for every $q\in\mathcal Q_\tau$.
Thus distinct physical falsifiers
can produce identical inference-invalidity
patterns even when compatibility sets differ.

\subsection{Unsupported Claims, Lost Valid Claims,
and Exact Inference Recovery}
\label{app:learned_compatibility_recovery}

We now characterize the inference consequences
of learning an approximate compatibility set.
Let $\Gamma$ be the true nonempty compatibility
set and let $\widehat{\Gamma}$ be a nonempty
estimated compatibility set, both contained
in the same physical state space.
Throughout this subsection, the prescribed
task family $\mathcal Q_\tau$ is finite.

\medskip
\noindent\textbf{Definition (Unsupported and
Lost Valid Claims).}
Define
\[
\begin{aligned}
\mathsf U_\tau(\Gamma,\widehat{\Gamma})
&=
V_\tau(\widehat{\Gamma})
\setminus V_\tau(\Gamma),\\
\mathsf L_\tau(\Gamma,\widehat{\Gamma})
&=
V_\tau(\Gamma)
\setminus V_\tau(\widehat{\Gamma}).
\end{aligned}
\]

The set $\mathsf U_\tau$ contains claims
declared valid by the estimated compatibility
set but not supported by the true compatibility
set. The set $\mathsf L_\tau$ contains claims
supported by the true compatibility set
but lost under the estimate.

These errors are task-relative: an unsupported
claim need not be false at the realized physical
state; it is unsupported because it does not
hold throughout the true compatibility set.

\medskip
\noindent\textbf{Definition (Task-Inference Error).}
The exact number of changed validity decisions is
\[
\begin{aligned}
E_\tau(\Gamma,\widehat{\Gamma})
&=
\left|
V_\tau(\Gamma)
\triangle
V_\tau(\widehat{\Gamma})
\right|\\
&=
|\mathsf U_\tau(\Gamma,\widehat{\Gamma})|
+
|\mathsf L_\tau(\Gamma,\widehat{\Gamma})|.
\end{aligned}
\]

This quantity measures disagreement in
the prescribed task claims rather than
geometric error between state sets.
Its magnitude depends on the task vocabulary:
semantically equivalent but separately listed
claims are counted separately, although
they induce the same inference-equivalence
distinction.

\medskip
\noindent\textbf{Theorem (Exact Characterization
of Inference Errors).}
For every prescribed claim
$q\in\mathcal Q_\tau$, let
\[
W_q(\Gamma)=\Gamma\cap T_q^c.
\]

Then
\[
\begin{aligned}
\mathsf U_\tau(\Gamma,\widehat{\Gamma})
=
\bigl\{q\in\mathcal Q_\tau:\,
&W_q(\Gamma)\neq\varnothing,\\
&W_q(\widehat{\Gamma})=\varnothing
\bigr\},
\end{aligned}
\]
and
\[
\begin{aligned}
\mathsf L_\tau(\Gamma,\widehat{\Gamma})
=
\bigl\{q\in\mathcal Q_\tau:\,
&W_q(\Gamma)=\varnothing,\\
&W_q(\widehat{\Gamma})\neq\varnothing
\bigr\}.
\end{aligned}
\]

Consequently,
\[
\begin{aligned}
E_\tau(\Gamma,\widehat{\Gamma})
=
\sum_{q\in\mathcal Q_\tau}
\Bigl|
&\mathbf 1\{W_q(\Gamma)=\varnothing\}\\
-
&\mathbf 1\{
W_q(\widehat{\Gamma})=\varnothing\}
\Bigr|.
\end{aligned}
\]

\noindent\textit{Proof.}
For any $q\in\mathcal Q_\tau$ and
$S\in\{\Gamma,\widehat{\Gamma}\}$,
\[
q\in V_\tau(S)
\iff W_q(S)=\varnothing.
\]
Substituting this equivalence into
the definitions of $\mathsf U_\tau$
and $\mathsf L_\tau$ gives both
claimed set identities.

These error sets are disjoint,
and their union is
$V_\tau(\Gamma)\triangle
V_\tau(\widehat{\Gamma})$.
Since $\mathcal Q_\tau$ is finite,
counting disagreements between
the binary validity indicators
gives the stated expression for $E_\tau$.

\medskip
\noindent\textbf{Theorem (Equivalent Conditions
for Exact Inference Recovery).}
For nonempty $\Gamma$ and
$\widehat{\Gamma}$, the following
conditions are equivalent:
\begin{enumerate}
\item
$E_\tau(\Gamma,\widehat{\Gamma})=0$;
\item
$\mathsf U_\tau(\Gamma,\widehat{\Gamma})
=\mathsf L_\tau(\Gamma,\widehat{\Gamma})
=\varnothing$;
\item
$V_\tau(\widehat{\Gamma})
=V_\tau(\Gamma)$;
\item
$\mathrm{Cl}_\tau(\widehat{\Gamma})
=\mathrm{Cl}_\tau(\Gamma)$;
\item
$F_\tau(\widehat{\Gamma})
=F_\tau(\Gamma)$.
\end{enumerate}

\noindent\textit{Proof.}
Since
$E_\tau=|\mathsf U_\tau|+|\mathsf L_\tau|$
and both terms are nonnegative,
(1) is equivalent to (2).
By the definitions of unsupported and
lost claims, (2) is equivalent to (3).
The task-closure characterization gives
(3)$\iff$(4), while exact-frontier
generation for finite $\mathcal Q_\tau$
gives (3)$\iff$(5).

\medskip
\noindent\textbf{Corollary (Asymmetry of
One-Sided Compatibility Errors).}
For nonempty compatibility sets:
\begin{enumerate}
\item
If $\Gamma\subseteq\widehat{\Gamma}$,
then
$\mathsf U_\tau(\Gamma,\widehat{\Gamma})
=\varnothing$.
\item
If $\widehat{\Gamma}\subseteq\Gamma$,
then
$\mathsf L_\tau(\Gamma,\widehat{\Gamma})
=\varnothing$.
\end{enumerate}

Neither inclusion alone guarantees
exact inference recovery.

\noindent\textit{Proof.}
If $\Gamma\subseteq\widehat{\Gamma}$,
antitonicity gives
$V_\tau(\widehat{\Gamma})
\subseteq V_\tau(\Gamma)$,
so $\mathsf U_\tau=\varnothing$.
If $\widehat{\Gamma}\subseteq\Gamma$,
the reverse inclusion gives
$\mathsf L_\tau=\varnothing$.

Neither inclusion guarantees equality:
take $\mathcal S_{\mathrm{phys}}=\{a,b\}$,
$\mathcal Q_\tau=\{q\}$, and $T_q=\{a\}$.
Then $q$ is valid for $\{a\}$ but not
for $\{a,b\}$. Assigning these two
sets to $\Gamma$ and $\widehat{\Gamma}$
in either order produces a lost or an
unsupported claim, respectively.

\medskip
\noindent\textbf{Example (Exact Inference
Without Exact Set Recovery).}
Let
\[
\mathcal S_{\mathrm{phys}}=\{a,b,c\},
\quad \mathcal Q_\tau=\{q\},
\quad T_q=\{a,b\}.
\]
Take $\Gamma=\{a\}$ and
$\widehat{\Gamma}=\{a,b\}$.
Although $\Gamma\neq\widehat{\Gamma}$,
both sets are contained in $T_q$.
Consequently,
\[
V_\tau(\Gamma)
=V_\tau(\widehat{\Gamma})=\{q\}.
\]
Thus $E_\tau(\Gamma,\widehat{\Gamma})=0$:
exact compatibility-set reconstruction
is sufficient but not necessary for
exact task-inference recovery.

\medskip
\noindent\textbf{Remark (Scope of Reliability).}
The preceding identities are deterministic
and distribution-free, relative to the
chosen true compatibility set and
prescribed task family.

For a registered finite-bank approximation,
the identities remain exact for that bank,
but this does not certify full-model
reliability. Such a guarantee additionally
requires a justified relation between
the bank and the full admissible
physical domain.

\subsection{Exact Inference Preservation
under Two-Sided Set Errors}
\label{app:two_sided_inference_preservation}

The preceding results characterize inference
errors induced by estimated compatibility sets.
We now establish an exact preservation criterion
that permits simultaneous removal and addition
of physical states.

\medskip
\noindent\textbf{Definition (Invalid-Claim Family).}
For a nonempty compatibility set $\Gamma$,
define
\[
\mathcal Q_\tau^-(\Gamma)
=
\mathcal Q_\tau\setminus V_\tau(\Gamma).
\]

Every claim in $\mathcal Q_\tau^-(\Gamma)$
has at least one falsifying state in $\Gamma$.

\medskip
\noindent\textbf{Theorem (Closure--Witness
Characterization).}
Let $\Gamma$ and $\widehat{\Gamma}$
be nonempty subsets of the physical state space.
Then
\[
V_\tau(\widehat{\Gamma})
=
V_\tau(\Gamma)
\]
if and only if both conditions hold:
\begin{enumerate}
\item
The estimated set remains inside the
task closure of the true set:
\[
\widehat{\Gamma}
\subseteq
\mathrm{Cl}_\tau(\Gamma).
\]
\item
Every originally invalid claim retains
at least one falsifying witness:
\[
\forall q\in\mathcal Q_\tau^-(\Gamma),
\quad
\widehat{\Gamma}\cap T_q^c
\neq\varnothing.
\]
\end{enumerate}

\noindent\textit{Proof.}
By the definition of task closure,
condition 1 is equivalent to
$\widehat{\Gamma}\subseteq T_q$
for every $q\in V_\tau(\Gamma)$,
and hence to
\[
V_\tau(\Gamma)
\subseteq V_\tau(\widehat{\Gamma}).
\]

For any $q\in\mathcal Q_\tau$,
\[
q\notin V_\tau(\widehat{\Gamma})
\iff
\widehat{\Gamma}\cap T_q^c
\neq\varnothing.
\]
Thus condition 2 is equivalent to
\[
V_\tau(\widehat{\Gamma})
\subseteq V_\tau(\Gamma).
\]

The two conditions therefore hold
simultaneously if and only if
$V_\tau(\widehat{\Gamma})
=V_\tau(\Gamma)$.

\medskip
\noindent\textbf{Definition (Falsifier
Witness Core).}
A subset $C_\Gamma\subseteq\Gamma$
is a falsifier witness core if
\[
\forall q\in\mathcal Q_\tau^-(\Gamma),
\quad
C_\Gamma\cap T_q^c\neq\varnothing.
\]

Such a core preserves at least one
counterexample to every claim that
is invalid under $\Gamma$.

\medskip
\noindent\textbf{Theorem (Finite Witness
Core and Preservation Corridor).}
Suppose $\mathcal Q_\tau$ is finite and
$\Gamma$ is nonempty.

There exists a finite falsifier witness core
$C_\Gamma\subseteq\Gamma$ satisfying
\[
|C_\Gamma|
\le
|\mathcal Q_\tau^-(\Gamma)|.
\]

Moreover, every nonempty set
$\widehat{\Gamma}$ satisfying
\[
C_\Gamma
\subseteq
\widehat{\Gamma}
\subseteq
\mathrm{Cl}_\tau(\Gamma)
\]
preserves exact task inference:
\[
V_\tau(\widehat{\Gamma})
=
V_\tau(\Gamma).
\]

\noindent\textit{Proof.}
For each $q\in\mathcal Q_\tau^-(\Gamma)$,
choose $s_q\in\Gamma\cap T_q^c$
and define
\[
C_\Gamma=
\{s_q:q\in\mathcal Q_\tau^-(\Gamma)\}.
\]
These choices exist by invalidity, and
$|C_\Gamma|\le|\mathcal Q_\tau^-(\Gamma)|$.
Thus $C_\Gamma$ is a finite
falsifier witness core.

For any nonempty $\widehat{\Gamma}$ with
$C_\Gamma\subseteq\widehat{\Gamma}
\subseteq\mathrm{Cl}_\tau(\Gamma)$,
the lower inclusion retains every
falsifying witness, while the upper
inclusion preserves every valid claim.
The closure--witness characterization
therefore gives
\[
V_\tau(\widehat{\Gamma})=V_\tau(\Gamma).
\]

If $\mathcal Q_\tau^-(\Gamma)=\varnothing$,
then $C_\Gamma=\varnothing$ and the
witness condition is vacuous.
The conclusion still holds for every
nonempty $\widehat{\Gamma}
\subseteq\mathrm{Cl}_\tau(\Gamma)$.

\medskip
\noindent\textbf{Corollary (Finite Exact-Inference
Witness Set).}
Under the preceding assumptions, every
nonempty compatibility set $\Gamma$ admits
a nonempty finite subset
$C_\Gamma^\star\subseteq\Gamma$ satisfying
\[
V_\tau(C_\Gamma^\star)=V_\tau(\Gamma),
\qquad
|C_\Gamma^\star|
\le\max\{1,|\mathcal Q_\tau^-(\Gamma)|\}.
\]

\noindent\textit{Proof.}
If $\mathcal Q_\tau^-(\Gamma)\neq\varnothing$,
take the witness core constructed in the
preceding theorem. It is nonempty and
preserves valid inference because
$C_\Gamma^\star\subseteq\Gamma
\subseteq\operatorname{Cl}_\tau(\Gamma)$.
Otherwise, choose any
$s\in\Gamma$ and set
$C_\Gamma^\star=\{s\}$.
Since every prescribed claim is valid
throughout $\Gamma$, all remain valid
on this singleton. This proves the claim.

\medskip
\noindent\textbf{Remark (Witness-Set Cardinality).}
The bound on $|C_\Gamma^\star|$ counts
physical witness states, not distinct
quotient code values, representation
dimensions, or storage bits.

\medskip
\noindent\textbf{Example (Simultaneous
State Deletion and Addition).}
Let
\[
\mathcal S_{\mathrm{phys}}
=
\{a,b,c,d,e\},
\]
with three task claims satisfying
\[
\begin{aligned}
T_{q_0}&=\{a,b,c,d\},\\
T_{q_1}&=\{a,c,d\},\\
T_{q_2}&=\{b,c,d\}.
\end{aligned}
\]

Take the true compatibility set
\[
\Gamma=\{a,b,c\}.
\]

Then
\[
V_\tau(\Gamma)=\{q_0\},
\qquad
\mathrm{Cl}_\tau(\Gamma)
=
\{a,b,c,d\}.
\]

The set
\[
C_\Gamma=\{a,b\}
\]
is a falsifier witness core:
$b$ falsifies $q_1$, while
$a$ falsifies $q_2$.

Consider the estimated set
\[
\widehat{\Gamma}=\{a,b,d\}.
\]

Relative to $\Gamma$, state $c$
has been removed and state $d$
has been added. Nevertheless,
\[
C_\Gamma
\subseteq
\widehat{\Gamma}
\subseteq
\mathrm{Cl}_\tau(\Gamma).
\]

Hence
\[
V_\tau(\widehat{\Gamma})
=
V_\tau(\Gamma)
=
\{q_0\}.
\]

Thus simultaneous state deletion
and addition need not alter any
validity decision.

\medskip
\noindent\textbf{Remark (Sufficiency
of a Fixed Witness Core).}
The closure--witness characterization
is necessary and sufficient, whereas
containment of a chosen $C_\Gamma$
is only sufficient: different falsifying
states may preserve the same invalid claims.

Constructing $C_\Gamma$ requires knowing
which claims are invalid under the true
compatibility set. The theorem does not
assume these witnesses are available
to a learning algorithm. Moreover,
a set within the preservation corridor
need not be realizable by the physical
observation operator.

\section{Additional Quotient-Code Statistics}
\label{app:quotient_statistics}

Table~\ref{tab:exp_quotient} reports the
exact quotient-code counts and the
cardinality-only comparison across all
four registered observation regimes.

\begin{table}[H]
\centering
\caption{
Quotient reduction on registered finite
state banks.
$A_{\mathrm{size}}$ is the optimal
in-sample exact-signature accuracy
using compatibility-set cardinality alone.
}
\label{tab:exp_quotient}
\setlength{\tabcolsep}{4pt}
\footnotesize
\begin{tabular}{lrrrr}
\toprule
Regime & $N$
& $N_\Gamma\!\to\!N_C$
& $R_{\mathrm{quot}}$
& $A_{\mathrm{size}}$ \\
\midrule
Historical
& 5,760 & $145\!\to\!31$
& 78.62\% & 47.85\% \\
Geometry-OOD
& 1,600 & $33\!\to\!22$
& 33.33\% & 41.12\% \\
Mechanism-OOD
& 800 & $33\!\to\!15$
& 54.55\% & 59.50\% \\
Joint-OOD
& 640 & $25\!\to\!18$
& 28.00\% & 36.25\% \\
\bottomrule
\end{tabular}
\end{table}

\begin{figure}[H]
\centering
\begin{tikzpicture}[
    x=1cm,
    y=1cm,
    scale=1.00,
    transform shape
]

\fill[blue!70!black]
    (0.12,0.60) rectangle (0.34,0.78);

\node[anchor=west,font=\scriptsize]
    at (0.44,0.69)
    {Required quotient codes};

\fill[orange!45]
    (4.10,0.60) rectangle (4.32,0.78);

\node[anchor=west,font=\scriptsize]
    at (4.42,0.69)
    {Reduced distinctions};

\foreach \yy/\regime/\nG/\nC/\reduction in {
    0/Historical/145/31/78.62,
    -0.87/Geometry-OOD/33/22/33.33,
    -1.74/Mechanism-OOD/33/15/54.55,
    -2.61/Joint-OOD/25/18/28.00
}{
    \pgfmathsetmacro{\ytop}{\yy+0.04}
    \pgfmathsetmacro{\ybottom}{\yy-0.16}
    \pgfmathsetmacro{\ylabel}{\yy+0.20}
    \pgfmathsetmacro{\ypercent}{\yy-0.14}

    \pgfmathsetmacro{\xcut}{
        2.05+4.05*(\nC/\nG)
    }

    \node[anchor=west,font=\scriptsize]
        at (0.05,\ylabel)
        {\regime};

    \fill[blue!70!black]
        (2.05,\ybottom)
        rectangle
        (\xcut,\ytop);

    \fill[orange!45]
        (\xcut,\ybottom)
        rectangle
        (6.10,\ytop);

    \draw[gray!65,thin]
        (2.05,\ybottom)
        rectangle
        (6.10,\ytop);

    \node[anchor=west,font=\scriptsize]
        at (6.30,\ylabel)
        {$\nG\rightarrow\nC$};

    \node[anchor=west,font=\scriptsize]
        at (6.30,\ypercent)
        {$\reduction\%$};
}

\end{tikzpicture}

\caption{
\textbf{Exact quotient-code reduction.}
Registered historical and OOD regimes
show the reduction from distinct
compatibility sets $N_\Gamma$ to
required quotient-code values $N_C$.
Blue denotes $N_C$ and orange denotes
$N_\Gamma-N_C$.
Labels report absolute counts and
reduction percentages.
The results concern distinct code values,
not storage or computational complexity.
}
\label{fig:quotient_reduction}
\end{figure}
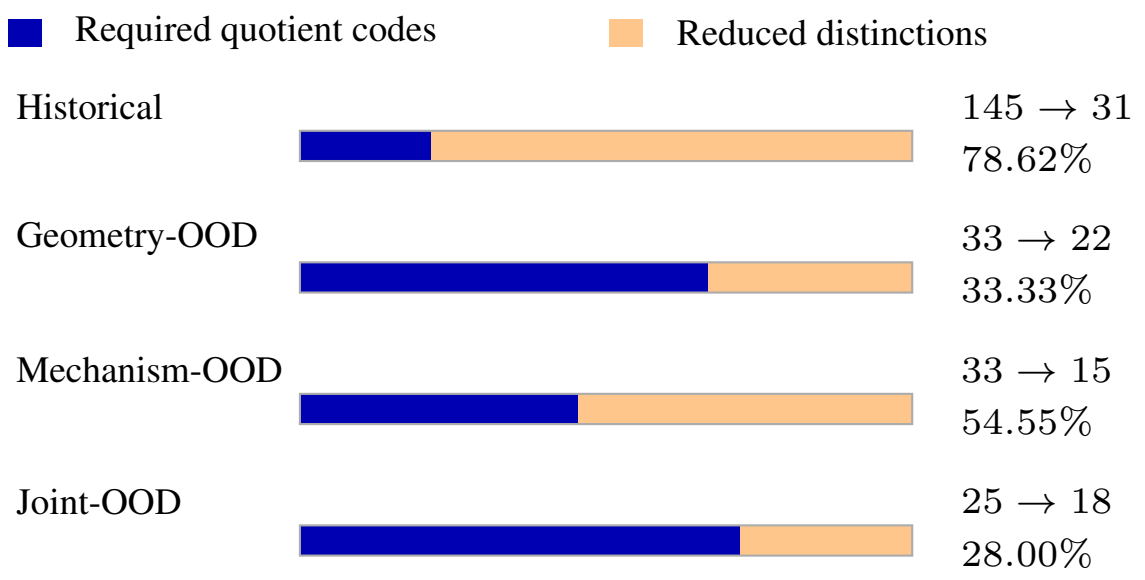


\section{Additional Observation-Ambiguity Statistics}
\label{app:observation_ambiguity_statistics}

Table~\ref{tab:supp_observation_frontier_results}
reports the complete ambiguity and
inference-frontier statistics across
the historical and three prospective
observation regimes.

\begin{table}[H]
\centering
\caption{
Physical ambiguity and valid inference on
registered banks. Hist.: historical;
Geo./Mech./Joint: prospective shifts.
All rates are percentages; True/unsupported
is normalized over true claim instances.
}
\label{tab:supp_observation_frontier_results}
\small
\setlength{\tabcolsep}{2.5pt}
\begin{tabular}{lrrrr}
\toprule
 & Hist. & Geo. & Mech. & Joint \\
\midrule
Observations
 & 5760 & 1600 & 800 & 640 \\
Ambiguous (\%)
 & 73.85 & 64.38 & 74.00 & 65.00 \\
Partial-ID (\%)
 & 49.38 & 41.00 & 55.75 & 48.44 \\
Distinct frontiers
 & 31 & 22 & 15 & 18 \\
Multi-maximal (\%)
 & 56.39 & 55.50 & 72.50 & 56.56 \\
True/unsupported (\%)
 & 37.13 & 31.04 & 28.16 & 28.42 \\
Pulse recovery (\%)
 & 67.80 & 88.44 & 68.75 & 88.28 \\
\bottomrule
\end{tabular}
\end{table}


\section{Official HRSID Test: Inference-Signature Evaluation}
\label{app:hrsid_test}

The real-SAR experiment in
Fig.~\ref{fig:official_test} evaluates
$N=1{,}962$ official HRSID test images
under seven registered SAR task claims.
Let $\widehat V_i^{(m)}$ denote the claim set
issued by method $m$. For visual 3-NN,
$\widehat V_i^{(\mathrm{3NN})}
=V_{\tau_{\mathrm{SAR}},\mathcal B}
(\widehat\Gamma_i^{(\mathrm{3NN})})$,
where the candidate set contains three
training-bank neighbors. Ridge and prevalence
provide claim-level controls with matched
per-image output counts. The reference signature $V_i^\star$ is
derived from the official COCO annotations.
The seven registered claims comprise three
ship-count thresholds
(\texttt{count\_ge\_2},
\texttt{count\_ge\_4},
\texttt{count\_ge\_8})
and four ship-center quadrant claims
(\texttt{ship\_center\_NW},
\texttt{ship\_center\_NE},
\texttt{ship\_center\_SW},
\texttt{ship\_center\_SE}).

For precise reference-label construction,
let $\mathcal A_i$ denote the official COCO
ship annotations of image $I_i$, whose
width and height are $W_i$ and $H_i$.
An annotated bounding box
$b=(x_b,y_b,w_b,h_b)$ has center
$c_b=(x_b+w_b/2,y_b+h_b/2)$,
using image coordinates with the origin
at the upper-left corner.
The three count claims are defined by
\begin{equation}
q_k^{\mathrm{count}}(I_i)
=\mathbf{1}\{|\mathcal A_i|\geq k\},
\qquad k\in\{2,4,8\}.
\end{equation}
The four spatial claims indicate whether
at least one annotated ship center lies
in the corresponding image quadrant.
These quadrants are
\begin{equation}
\begin{aligned}
R_{\mathrm{NW}}^i
&=\{(u,v):u<W_i/2,\ v<H_i/2\},\\
R_{\mathrm{NE}}^i
&=\{(u,v):u\geq W_i/2,\ v<H_i/2\},\\
R_{\mathrm{SW}}^i
&=\{(u,v):u<W_i/2,\ v\geq H_i/2\},\\
R_{\mathrm{SE}}^i
&=\{(u,v):u\geq W_i/2,\ v\geq H_i/2\}.
\end{aligned}
\end{equation}
For $r\in\{\mathrm{NW},\mathrm{NE},
\mathrm{SW},\mathrm{SE}\}$, define
\begin{equation}
q_r^{\mathrm{quad}}(I_i)
=\mathbf{1}\left\{
\exists b\in\mathcal A_i:
c_b\in R_r^i
\right\}.
\end{equation}
Thus, centers on the vertical and
horizontal image midlines are assigned
to the eastern and southern halves,
respectively. Multiple quadrant claims
may be true for the same image.
An image with no annotated ships
satisfies none of the seven claims.

Let $\mathcal Q_{\mathrm{SAR}}$ denote
this fixed seven-claim family.
The annotation-relative reference
signature is
\begin{equation}
V_i^\star=
\{q\in\mathcal Q_{\mathrm{SAR}}:
q(I_i)=1\}.
\end{equation}

We independently reconstructed the
reference signatures from the official
HRSID COCO annotations. All $1{,}962$
test images were matched successfully,
and the $5{,}922$ annotated ship boxes
reproduced the recorded ship counts
without discrepancy. The three count
claims agreed with the frozen reference
labels for every image. The four spatial
claims were also reconstructed exactly,
including cases with ship centers on
the image midlines. Among four tested
axis-wise midpoint-assignment conventions,
the convention above was the only one
with zero signature mismatches.
This post-freeze consistency audit
did not alter the trained models,
frozen predictions, or evaluation protocol.
The reference signatures are relative
to the official annotations and do not
constitute certificates of complete
physical-scene reconstruction.

The registered candidate pool contains
$3{,}642$ official HRSID training images.
Frozen visual features concatenate a
$32\times32$ box-resampled grayscale image,
a $16\times16$ max-pooled representation
obtained from $64\times64$ bilinear resizing,
and six intensity statistics, yielding
$1{,}286$ dimensions. Within each of five
grouped folds, features are standardized
and projected onto 32 principal components
using training-fold data only. Visual 3-NN
retrieves the three nearest training
candidates by Euclidean distance and
issues a claim only when all three
candidates support it.

The frozen protocol uses source-prefix
grouping and exact-duplicate connectivity
controls to constrain the training folds.
Official test annotations are accessed
only after the test predictions have been
frozen. Source-prefix grouping serves as
a proxy for source separation and does not
certify independence of physical SAR scenes.

Define correctly issued claims
$T_m=\sum_i|\widehat V_i^{(m)}\cap V_i^\star|$
and total issued claims
$I_m=\sum_i|\widehat V_i^{(m)}|$.
Claim precision and correctly supported
claims per image are $T_m/I_m$ and $T_m/N$,
respectively; exact task-signature recovery is
$N^{-1}\sum_i\mathbf{1}
[\widehat V_i^{(m)}=V_i^\star]$.

Across the $1{,}962$ official HRSID test
images, visual 3-NN produces $1{,}913$
distinct candidate sets but only $36$
distinct inference signatures over the
seven registered claims. The empirical
task-relative quotient therefore has
$98.12\%$ fewer distinct code values
($36$ rather than $1{,}913$) on this
evaluated family, while preserving
the issued inference signature of
every candidate set. This reduction
concerns distinguishable task codes,
not representation dimensionality,
storage, or computational cost.

\begin{table}[t]
\centering
\caption{Inference-signature evaluation on the
official HRSID test set ($N=1{,}962$).
All methods issue $1{,}347$ claims with
matched per-image output counts.
$T_m$ is the number of correctly supported
claims; precision and exact recovery
are reported as percentages.}
\label{tab:hrs_sar_claim_results}
\footnotesize
\setlength{\tabcolsep}{3pt}
\begin{tabular}{lrrrr}
\hline
Method & $T_m$ & Prec. & $T_m/N$ & Exact \\
\hline
Visual 3-NN & 1,176 & 87.31 & 0.599 & 23.55 \\
Ridge & 771 & 57.24 & 0.393 & 3.21 \\
Prevalence & 700 & 51.97 & 0.357 & 0.10 \\
\hline
\end{tabular}
\end{table}

Table~\ref{tab:hrs_sar_claim_results}
reports the absolute inference results
under matched per-image claim budgets. Paired source-prefix
cluster bootstrap (2,000 replicates
across 136 groups) shows that visual
3-NN increases correctly supported
claims per image by $0.2064$
(95\% CI $[0.1778,0.2332]$) over ridge
and $0.2426$ (95\% CI
$[0.2094,0.2734]$) over prevalence.

Claim precision evaluates the correctness
of issued claims, whereas exact-signature
recovery requires agreement with the
reference on every registered claim,
including those not issued by the method.
Consequently, high claim precision does
not imply complete recovery of the
reference inference signature. These
metrics distinguish the reliability of
individual issued conclusions from the
completeness of task-level inference.

The witnesses in Fig.~\ref{fig:official_test}
were selected after evaluation.
Results are bank- and claim-relative;
they neither certify physical reconstruction
nor isolate quotient-coding gains from
candidate-set construction.

A separate reproducibility code archive reproduces
the controlled-bank frozen evaluation and includes
HRSID algorithm self-tests. This archive is not
included with the present arXiv preprint.
The official HRSID numerical results were separately
verified against author-retained frozen prediction
and evaluation records, which are not included in
the code archive. Consequently, the archive alone
does not reproduce the official HRSID numerical
evaluation, upstream model training, image-feature
extraction, or prediction generation from raw SAR
images.

The real-SAR inference-equivalence results
are restricted to the registered seven-claim
family and the observed candidate sets.
The reduction in distinct code values does not
establish a reduction in computational cost or
an improvement in predictive accuracy caused
by quotient coding.

\end{document}